\documentclass[10pt,twocolumn,letterpaper]{article}

\usepackage{iccv}
\usepackage{times}
\usepackage{epsfig}
\usepackage{graphicx}
\usepackage{amsmath}
\usepackage{amssymb}

\usepackage{amsmath}
\usepackage{amssymb}
\usepackage{booktabs}
\usepackage{microtype}
\usepackage{enumitem}
\usepackage{xcolor}

\usepackage{comment}
\usepackage{caption}
\usepackage{subcaption}

\usepackage{amsmath}
\DeclareMathOperator{\arctantwo}{arctan2}

\usepackage{multicol}

\usepackage{gensymb}

\usepackage{authblk}

\newcommand{\jason}[1]{\textcolor{blue}{[Jason: #1]}}

\usepackage[pagebackref=true,breaklinks=true,letterpaper=true,colorlinks,bookmarks=false]{hyperref}

\usepackage[capitalize]{cleveref}
\crefname{section}{Sec.}{Secs.}
\Crefname{section}{Section}{Sections}
\Crefname{table}{Table}{Tables}
\crefname{table}{Tab.}{Tabs.}

\iccvfinalcopy 

\def\iccvPaperID{6795} 

\ificcvfinal\fi

\begin{document}

\title{Conditional 360-degree Image Synthesis for Immersive Indoor Scene Decoration}

\makeatletter
\renewcommand\AB@affilsepx{ \protect\Affilfont}
\makeatother

\author[1]{Ka Chun Shum}
\author[1]{Hong-Wing Pang}
\author[2]{Binh-Son Hua}
\author[3]{Duc Thanh Nguyen}
\author[1]{Sai-Kit Yeung}

\affil[1]{Hong Kong University of Science and Technology}
\affil[2]{VinAI Research, Vietnam}
\affil[3]{Deakin University}

\maketitle
\ificcvfinal\thispagestyle{empty}\fi

\begin{abstract}
   
   
   In this paper, we address the problem of conditional scene decoration for 360\degree images. Our method takes a 360\degree background photograph of an indoor scene and generates decorated images of the same scene in the panorama view. To do this, we develop a 360-aware object layout generator that learns latent object vectors in the 360\degree view to enable a variety of furniture arrangements for an input 360\degree background image. We use this object layout to condition a generative adversarial network to synthesize images of an input scene. 
   To further reinforce the generation capability of our model, we develop a simple yet effective scene emptier that removes the generated furniture and produces an emptied scene for our model to learn a cyclic constraint. We train the model on the Structure3D dataset and show that our model can generate diverse decorations with controllable object layout. Our method achieves state-of-the-art performance on the Structure3D dataset and generalizes well to the Zillow indoor scene dataset. Our user study confirms the immersive experiences provided by the realistic image quality and furniture layout in our generation results. Our implementation will be made available.
\end{abstract}

\section{Introduction}
\label{sec:intro}

Panoramas (360\degree images) enable immersive user experiences and have been applied intensively to various virtual reality (VR) applications~\cite{aitamurto2018sense,brivio2021virtual,ritter2022three}. However, automated generation of indoor scenes in the 360\degree view for architectural and interior design remains understudied due to many challenges. First, the generation process must conform the common distortions in the 360\degree view. Second, generated content must be controllable.


Common generative models, e.g., StyleGAN~\cite{karras2019stylegan,karras2020stylegan2} can generate photorealistic images. However, these methods are unconditional generation techniques, i.e., an output image is generated from a random code sampled in a latent space without interpreted meaning, thus limiting content controllability. Existing conditional image synthesis techniques, e.g., image-to-image translation~\cite{isola2017image,zhu2017toward,wang2018high,tang2019multi}, on the other hand, do not have explicit support for scene representations and thus have limited capability for scene manipulation.


In this work, we focus on conditional image synthesis of 360\degree indoor scenes. We are inspired by the neural scene decoration (NSD) in~\cite{pang2022nsd}, aiming to generate a decorated scene image from a given background image and user-defined furniture arrangement. However, the NSD method in~\cite{pang2022nsd} has several limitations. First, it requires an object layout modeling furniture arrangement from users, making the generation process not fully automatic. Second, its object layout, represented by rectangles, is not applicable in the 360\degree view using equirectangular projection~\cite{snyder1997flattening}. Third, there is no mechanism to control different attributes of the generated furniture, limiting the diversity of the generated content. 



We instead take a different approach for scene representation and propose a conditional image synthesis method for automatic scene decoration in the 360\degree setting. 
We first develop a 360-aware object layout generator that learns a set of object vectors representing the furniture arrangement of a 360\degree scene. We use this layout as the latent representation in a generative adversarial network to condition the generated content. To support the training of the layout and generative adversarial network, we devise a scene emptier that performs a dual task, i.e., making a decorated scene empty. 
In summary, we make the following contributions in our work.
\begin{itemize}[leftmargin=*]
    \item A 360-aware object layout generator that automatically learns an object arrangement from a 360\degree background image. Generated layouts condition the scene decoration in the 360\degree viewer;
    \item A novel generative adversarial network (GAN) to synthesize diverse and controllable scene decorations in the 360\degree setting;
    \item A scene emptier for reinforcement of the conditioning ability and generation ability in the training;
    \item Extensive experiments and user studies on benchmark datasets including the Structured3D~\cite{zheng2020structured3d} and Zillow Indoor dataset~\cite{ZInD} to validate our method and to provide immersive experiences to users.
\end{itemize}



\begin{figure*}
  \centering
    \includegraphics[width=\linewidth]{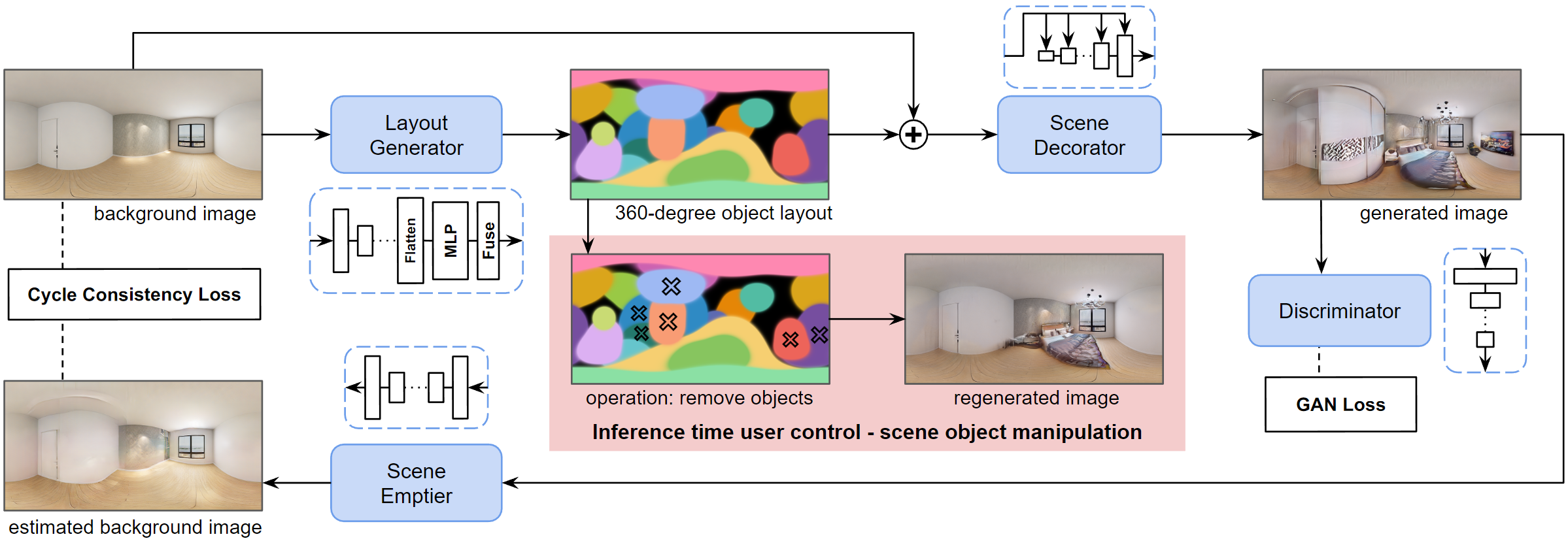}
    \vspace{-0.1in}
    \caption{Overview of our method. Input is a 360\degree background image of an empty scene. The input is fed to a layout generator to produce a set of object vectors to form a 360\degree object layout. The object layout and input background image are integrated to condition a GAN architecture (our decorator and discriminator) to generate a decorated image of the same scene. During training, the output decorated image is fed to a scene emptier to render back the background image of the empty scene. This estimated background is compared with the input background for a cyclic constraint. At inference time, users can manipulate the object vectors to produce different object layouts to generate diverse images.}
    \label{fig:network_sturcture}
\end{figure*}

\section{Related work}
\label{sec:Related works}

\noindent\textbf{Neural image synthesis.} Existing neural image synthesis techniques can be grouped in two main directions: image-to-image translation and generative adversarial neural networks (GANs). Image-to-image translation methods~\cite{isola2017image,zhu2017toward,wang2018high,tang2019multi} aim at translating images from one domain to another. Among these, CycleGAN~\cite{zhu2017unpaired} with a cycle-consistency loss is well-known for its robustness yet effectiveness due to not requiring image pairs in both domains for training. Recent methods such as SPADE~\cite{park2019SPADE} and OASIS~\cite{schonfeld2021oasis} translate semantic maps into realistic images. We do not use semantic maps in our work because semantic annotation of 360\degree images is a costly task; drawing object silhouettes in a semantic map is also complex for novice users. Another difficulty for automatic decoration of 360\degree images is the difference in the object arrangement between the input and output image, making the translation challenging to pixel-level image translation methods.


Recent developments in GANs have sparked great interest in image synthesis, e.g., the family of StyleGAN~\cite{karras2019stylegan,karras2020stylegan2,karras2020ada,karras2021alias}. These models have demonstrated groundbreaking results in generating human faces~\cite{karras2019stylegan} and on some in-the-wild datasets~\cite{brock2018large}. They can also be conditioned on layouts for image synthesis~\cite{li2019layoutgan,yang2019semantic}. Several methods improve the quality of generated images using various cues such as layout reconfiguration~\cite{sun2019reconfig}, object context~\cite{he2021context}, and locality~\cite{li2021locality}.
For indoor scene image synthesis, 
ArchiGAN~\cite{chaillou2019archigan} and
HouseGAN~\cite{nauata2020house} generate apartment rooms and furniture layouts. BachGAN~\cite{li2020BachGAN} hallucinates a background from an object layout. 
NSD~\cite{pang2022nsd} conditions an image generator on both a background image and an object layout defined by users. Our method is perhaps the most related to~\cite{pang2022nsd} in the problem setting, but we address a more challenging problem where the object layout is learned automatically, eliminating the need for user input while enabling controllability in the generated content.


\noindent\textbf{360-degree image synthesis.} Several methods employ generators that produce smaller spatially-aware patches, which can be assembled together into a high-resolution, seamless output image. For example, COCO-GAN~\cite{lin2019coco} synthesizes a cylindrical set of patches to be assembled into a 360\degree panorama. InfinityGAN~\cite{lin2022infinitygan} generates in-between patches between two fixed patches via the latent code inversion procedure in~\cite{cheng2022inout}. Several works show that a panorama can be synthesized from various conditional information, such as from a single perspective image~\cite{akimoto2022diverse}, multiple perspective images~\cite{sumantri2020360} or aerial views~\cite{wu2022cross}.

\noindent\textbf{Indoor scene modeling.}
Traditional indoor scene modeling methods reason the 3D space, with analysis on structural and functional aspects of the space, for furniture arrangement. Early attempts include creating a physical model of a scene for object insertion~\cite{karsch2011legacy,FisherRSFH12,karsch2014composite,karsch2015thesis}, optimizing the spatial arrangement of furniture~\cite{Germer2009,yu2011make} with additional consideration of object relations and room attributes~\cite{henderson2017automatic,LiPXCKSTCCZ19}, and spatial constraints such as relation graph prior~\cite{wang2019planit,hu2020graph2plan,nauata2020house} and convolution prior~\cite{wang2018deep}.
Recently, Ritchie et al.~\cite{ritchie2019fast} used neural networks to predict the category, location, orientation, and dimension of objects in a top-down view. Zhang et al.~\cite{zhang2020deep} optimized a GAN-based architecture that models object position and orientation, where the discriminator takes both rendered images and 3D shapes into account. 
Compared with existing scene unfurnishing~\cite{zhang2016emptying} and scene furnishing nethods~\cite{yu2015clutter,zhang2021mageadd,liang2021decorin}, our method is image-based and thus does not require the use of 3D models. 

\section{Proposed Method}
\label{sec:Methodology}

\begin{figure*}[t!]
\centering
    \begin{subfigure}[t]{0.36\textwidth}
    \centering
    \includegraphics[width=\textwidth]{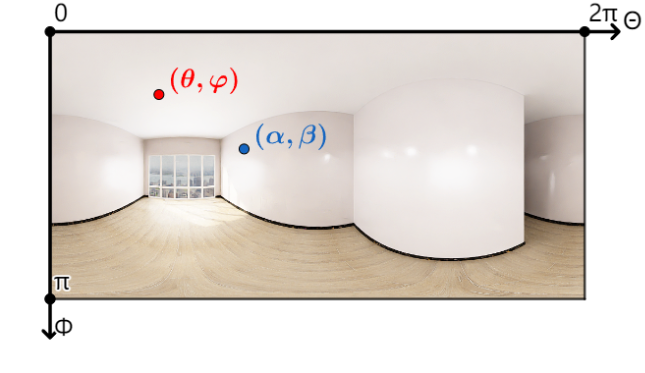}
    \caption{}
    \label{fig:model:pixel2ellipse_distance_calculation:pano}
\end{subfigure}\hfill
\begin{subfigure}[t]{0.32\textwidth}
    \centering
    \includegraphics[width=\textwidth]{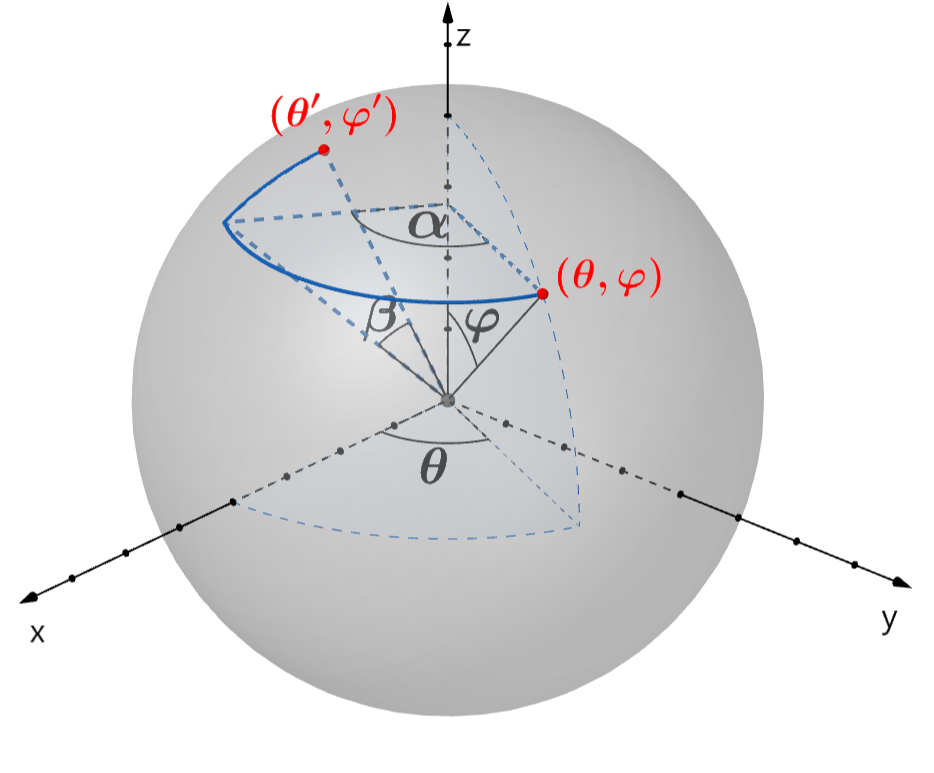}
    \caption{}
    \label{fig:model:pixel2ellipse_distance_calculation:sphere}
\end{subfigure}\hfill
 \begin{subfigure}[t]{0.28\textwidth}
    \centering
    \includegraphics[width=\textwidth]{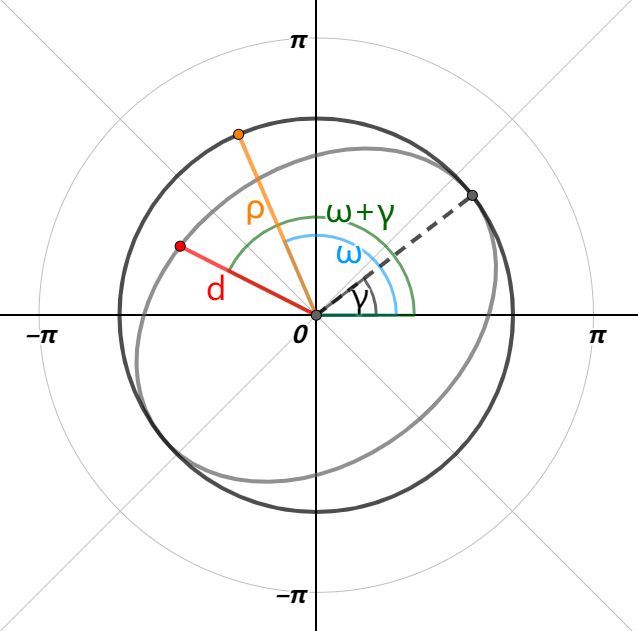}
    \caption{}
    \label{fig:model:pixel2ellipse_distance_calculation:ellipse}
\end{subfigure}
\caption{Visualization of calculating the distance $d$ from a pixel point $(\theta, \phi)$ to an ellipse object center $(\alpha, \beta)$. A panorama image (a) is modeled in the spherical coordinate system (b) and then rotated to align with the image origin to $(\alpha, \beta)$ in (b). In (c), the rotated image is projected to the polar coordinate system to effectively model an ellipse given ellipse rotation $\gamma$ and eccentricity $e$.}
\label{fig:model:pixel2ellipse_distance_calculation}
\end{figure*}

We propose a conditional model for automatic scene decoration for 360\degree images. Given a 360\degree background image $X$ that captures an empty scene, our model generates a 360\degree image $\hat{Y}$ of the scene in $X$, but with furniture. We use the equirectangular format to represent 360\degree images, where each pixel's $x$ and $y$ coordinate are mapped to the azimuth and polar angle in a spherical coordinate system, respectively. Our model has three sub-modules: (1) a conditional layout generator, (2) a conditional scene decorator (a GAN architecture), and (3) a scene emptier. The layout generator, trained in an unsupervised manner, disentangles possible objects to be generated in $X$ into an object layout $L$ that uses a set of latent vectors to represent objects in the 360\degree setting. 
The decorator generates $\hat{Y}$ by conditioning on the background image $X$ and the predicted object layout $L$.
The scene emptier clears up the decorated image $\hat{Y}$ to revert it to the input background image. The scene emptier is used in training of our model to reinforce its conditioning and generation ability via a cycle loss. We illustrate our method in Figure~\ref{fig:network_sturcture} and describe its sub-modules in the corresponding sub-sections.

\subsection{Conditional 360-aware object layout generator}
\label{sec:layout}

Our aim is to estimate and represent a possible furniture arrangement from the given background image $X$ in the 360\degree setting.
Our layout generator is a conditional image encoder followed by a multi-layer perceptron (MLP) to map the background image $X$ into a proper set of object vectors in the 360\degree view. Moreover, rather than representing the set of objects in a 2D plane~\cite{epstein2022blobgan}, our layout generator considers distortions and left-right boundary discontinuity artifacts in the omnidirectional view. Mathematically, we let each object vector composed by an ellipse location $\alpha, \beta \in \mathbb{R}$, an ellipse size $s \in \mathbb{R}$, an ellipse rotation $\gamma \in \mathbb{R}$, an ellipse eccentricity $e \in [0,1)$, and a feature vector $f \in \mathbb{R}^{d_f}$. 
The layout generator can be defined by a function that maps $X \in \mathbb{R}^{W \times H \times 3} \mapsto \{(\alpha_i, \beta_i, s_i, \gamma_i, e_i, f_i)\}_{i=1}^n \in \mathbb{R}^{n \times (5+d_f)}$
where $n$ is the number of object ellipses to generate, $W$ and $H$ are respectively the width and height of the image $X$.

To make the object vectors adaptive to a GAN architecture, we reshape them into an object layout $L \in \mathbb{R}^{W \times H \times d}$ in the same spatial dimension with $X$. Intuitively, we assume that a pixel closer to an object ellipse should convey more information about that ellipse. This can be modeled by measuring the distance $d$ from each pixel $(\theta, \phi) \in {\{(0, 2\pi], [0, \pi]\}}$ to every ellipse center $(\alpha, \beta)$, where the tuple $(\theta, \phi)$ is the sphere coordinate of a pixel in a 360\degree image. 
Calculating the distance $d$ requires geometric manipulations. An option is to use geodesic distance on a sphere and model the object as a circle instead of an ellipse. However, we found this results in collapsed object size during training possibly due to difficulty in modeling of irregular-shaped objects. 

Instead, we propose the following distance calculation, which is visualized in \cref{fig:model:pixel2ellipse_distance_calculation}. We summarize the main steps in calculating $d$ as follows. 
First, we align the sphere image center with the ellipse location $(\alpha, \beta)$ by rotating the sphere with the right-hand rule to obtain a rotated coordinate $(\theta', \phi')$. Next, we project the sphere image specified by $(\theta', \phi')$ to 2D polar coordinate system $(\rho, \omega)$. Finally, we count the effect of ellipse rotation $\gamma$ by adding it to polar coordinate $\omega$ and shrink the shape of the ellipse with ellipse eccentricity $e$ to get the final distance $d$.
We refer the readers to our supplementary material for detailed equations.

%

%

%
After calculating distance $d$, we fuse the features $f$ based on the inverse of $d$ and ellipse size $s$ to make a feature opacity $o=\mathrm{sigmoid}(s-d)$ for each ellipse. The feature vector at a location $(\theta, \phi)$ in the object layout $L$ is computed using alpha-compositing~\cite{epstein2022blobgan, porter1984compositing}:
\begin{align}
    L(\theta, \phi)
    =
    \sum\limits_{i}^n\{f_{i}o_{i}\prod\limits_{k=i+1}^n (1 - o_{k})\}.
    \label{eq:distance:opacity_and_alphacomposition}
\end{align}


\subsection{Conditional scene decorator}
\label{subsec:decorator}

We adopt the generator $G$ and the discriminator $D$ from StyleGAN2~\cite{karras2020stylegan2} for our conditional scene decorator. The input of the decorator includes the background image $X$ with the object layout $L$. Like~\cite{park2019semantic}, we split $L\in \mathbb{R}^{W \times H \times d_{f}}$ into $L_{u} \in \mathbb{R}^{W \times H \times d_{u}}$ and $L_{y} \in \mathbb{R}^{W \times H \times d_{y}}$ where $d_f = d_{u} + d_{y}$. $L_u$ and $L_y$ capture the structure and style information of the input scene, respectively. These maps are input for the generator $G$ where $L_{u}$ is considered for convolution operations and $L_{y}$ is considered for spatial modulation~\cite{park2019semantic}. 


To further strengthen the conditioning ability on $X$ and preserve its high-frequency information, we concatenate $L_{u}$ and $X$ and pass the concatenated result to $G$ pyramidally. The output of $G$ is a synthetically decorated image $\hat{Y}$, which is then classified (as real vs. fake) by the discriminator $D$.

\subsection{Scene emptier}
\label{subsec:emptier}
Ideally, removing decorated objects from the image $\hat{Y}$ should result in the background $X$. We apply this duality to reinforce the generation quality of our model. Specifically, we create a scene emptier $E$ that transforms a decorated image of a scene into an empty version of that scene. The emptier is implemented as an encoder-decoder architecture (see our supplementary material). We pretrain $E$ together with an unmodified version of the discriminator from StyleGAN2~\cite{karras2020stylegan2}, denoted as $D_{emp}$, using the following losses:
\begin{align}
    \mathcal{L}_{G_{emp}} &= \mathbb{E}_Y[1-D_{emp}(E(Y))],  \\
    \mathcal{L}_{D_{emp}} &= 
    \mathbb{E}_X[1-D_{emp}(X)]
    + \mathbb{E}_Y[D_{emp}(E(Y))], \\
    \mathcal{L}_{recon} &= \| X - E(Y) \|_2^2, \\
    \mathcal{L}_{emp} &= \mathcal{L}_{G_{emp}} + \mathcal{L}_{D_{emp}} + \mathcal{L}_{recon},
  \label{eq:emptierloss}
\end{align}
where $Y$ and $X$ represent a ground-truth decorated image and an empty image from training data. 

Given the decorated image $\hat{Y}$, the produced background $E(\hat{Y})$ from the pretrained scene emptier is used to form a cycle consistency loss between $E(\hat{Y})$ and $X$ to train the scene decorator. We note that the scene emptier and the cyclic constraint are necessary for the conditioning ability and generation ability of our model. This is because scene decoration is a weakly-constrained problem as there could be multiple solutions given a single background. Therefore, directly comparing the generated content $\hat{Y}$ with its ground-truth $Y$ via pairwise losses (MSE, perceptual loss) would hinder the diversity of the synthesis since there is only one ground-truth decoration per input image. The scene emptier, with cycle-consistency loss, can relax the hardness of the pairwise losses while enforcing the background consistency.


We emphasize that the selections of the architecture for $E$ and $D_{emp}$ are not of significance as the decorated-to-empty translation task is simpler than the empty-to-decorated translation task, which rigorously requires reasonable object arrangements. This observation allows us to choose a simple design for the emptier. As shown in experimental results, a simple emptier already suffices to strengthen the entire scene decoration process. 

We opt for pretraining the scene emptier before training the scene decorator as it leads to improved generation quality with the cycle consistency loss being a critic. 
This is explained by that with pretraining, the emptier is trained only with ground-truth decorated images and so it implicitly boosts the decorator to generate ground-truth-like results to fit the cycle consistency loss. In contrast, when the emptier is jointly trained with the scene decorator from scratch, the emptier can learn to empty low-quality decorated images synthesized in early iterations, and eventually tolerates such low-quality images in the learning process, degrading the overall performance of the entire pipeline.

\subsection{Horizontal circular padding}
A typical property of a panorama image is that the left and right boundaries loop around. However, convolutional layers in a neural network are weak in capturing information across the left-right boundaries of panorama images. Like~\cite{schubert2019circular}, we overcome this issue by applying circular padding. Precisely, for all the convolutional layers in our networks ($L$, $G$, $D$), we circularly pad pixels from the left to the right boundary and vice versa prior to performing convolutions, while regular padding is applied to the top and bottom boundaries. 

\subsection{Training objectives}
Given the pretrained emptier $E$, we train our entire model by a loss $\mathcal{L}_{total}$:
\begin{equation}
    \mathcal{L}_{total} = {\lambda}_{GAN}(\mathcal{L}_{G}
    + \mathcal{L}_{D}) + {\lambda}_{cycle} \mathcal{L}_{cycle},
  \label{eq:loss_total}
\end{equation}
which includes GAN losses ($\mathcal{L}_{G}$, $\mathcal{L}_{D}$) and a cycle loss $\mathcal{L}_{cycle}$ that leverages the emptier $E$ to impose a cyclic constraint on the background image $X$; $\lambda_{GAN}$ and $\lambda_{cycle}$ are the coefficients of the corresponding losses, respectively.

The losses $\mathcal{L}_{G}$ and $\mathcal{L}_{D}$ are defined as: 
\begin{align}
    \mathcal{L}_{G} &= \mathbb{E}_{\hat{Y}}[1-D(\hat{Y})], \\
    \mathcal{L}_{D} &= \mathbb{E}_{\hat{Y}}[D(\hat{Y})]
    +
    \mathbb{E}_Y[1-D(Y)],
  \label{eq:ganloss}
\end{align}
where $Y$ is a decorated image from the ground truth.

The cycle loss $\mathcal{L}_{cycle}$ constrains the consistency of the background image $X$ and the empty version  $E(\hat{Y})$ made by the emptier $E$ via a reconstruction loss:
\begin{equation}
    \mathcal{L}_{cycle} = \| X - E(\hat{Y}) \|_2^2.
  \label{eq:loss_cycle}
\end{equation}

\section{Experiments}
\label{sec:Experiments}

\subsection{Dataset}
We trained and evaluated our method on the Structured3D dataset~\cite{zheng2020structured3d}. To the best of our knowledge, it is the only dataset that contains a significant amount of paired unfurnished and furnished 360\degree images. The Structured3D dataset provides 21,835 360\degree image pairs rendered from distinct rooms in 3,500 indoor scenes. We trained our method and report its performance only on the bedroom subset and living room subset of the Structured3D dataset since only these two sets contain a sufficient number of images for training. We split the bedroom subset into 3,318 training and 350 test images, and the livingroom subset into 1,900 training and 237 test images. We also tested our model on the test set of the Zillow Indoor Dataset (ZInD)~\cite{ZInD}, which consists of 4,359 undecorated 360\degree images.

To increase the scale of the training data (for the bedroom subset), we applied panoramic-specific data augmentation. Particularly, except for random horizontal flipping, we implemented random horizontal circular translation on panorama images. Since the content crossing the left-right boundaries of a panorama image is connected, we circularly padded a random number of columns of pixels at the left to the right boundary to construct more panorama images. 


\subsection{Baselines}
Our primary goal is to synthesize a decorated 360\degree image given an unfurnished 360\degree image and to provide a certain level of object control. This task could be partially tackled by conditional image-to-image (I2I) translation methods as they translate images to a target domain although they do not provide controllability over the generated objects. Therefore, we compare our method with well-known and state-of-the-art I2I works including Pix2PixHD~\cite{wang2018high} that uses a one-to-one paired reconstruction loss to model domain translation, StarGANv2~\cite{choi2020stargan} that learns a one-to-many image translation model, and StyleD~\cite{kim2022style} that learns to implicitly categorize images in the target image domain and provide translation control towards categorized image domain.

For conditional layout-based generation methods, we compare our work with the methods by Pang et al.~\cite{pang2022nsd} and He et al.~\cite{he2021context} which achieve state-of-the-art performance in conditional image synthesis for scene decoration. As these methods additionally require ground-truth object labels and bounding boxes (not used in our model), to adapt them to our task, following~\cite{pang2022nsd}, we generate object layouts by extracting object bounding boxes from semantic and instance maps from the ground-truth of the Structured3D dataset.


\subsection{Implementation details}

We present implementation details of our layout generator, decorator, and emptier in the supplementary material. We set the number of object ellipses $n$ to 20 and the feature dimension $d_{f}$ to 1024. The emptier and the entire model are trained using the Adam optimizer~\cite{kingma2014adam} with a learning rate of 0.01. We set ${\lambda}_{GAN} = 1$ and ${\lambda}_{cycle} = 5$. The model was trained on equirectangular images. However, since several baselines require square images for training, we reshaped rectangular images into square images in both training and testing. Particularly, we experimented with our methods and other baselines in~\cref{sec:experimental} under $512\times512$ resolution and ablation study in~\cref{sec:ablation} under $256\times256$ resolution.

\subsection{Results}
\label{sec:experimental}
\noindent\textbf{Quantitative results.}
We quantitatively evaluate our method and compare it with other baselines using the Frechet Inception Distance (FID)~\cite{heusel2017gans} and Kernel Inception Distance (KID)~\cite{binkowski2018demystifying} metrics. FID and KID assess the generation quality of a method by measuring the similarity (in feature space) between images generated by that method and those from the ground-truth. We use KID$\times10^3$ in all experiments. 

As reported in Table~\ref{tab:quantitative_results}, our method outperforms all the baselines on both FID and KID scores. The conditional layout-based methods generally perform better than I2I methods except for the Pix2PixHD~\cite{wang2018high}. We speculate the reason is that layout-based methods receive extra hints from explicit object layout to better model object distribution. Meanwhile, I2I methods commonly have difficulty in object understanding, except for the Pix2PixHD~\cite{wang2018high} that uses a one-to-one paired loss (to ground truth decorated images). 


\begin{table}[]
    \centering
    \begin{tabular}{l|cc|cc}
        \toprule
         & \multicolumn{2}{c|}{bedroom} & \multicolumn{2}{c}{living room} \\
        \midrule
        Method & $\mathrm{FID} \downarrow$ & $\mathrm{KID} \downarrow$ & $\mathrm{FID} \downarrow$ & $\mathrm{KID} \downarrow$ \\
        \midrule
        Pix2PixHD~\cite{wang2018high} & 73.33 & 20.56 & 83.64 & 14.20\\ 
        StarGANv2~\cite{choi2020stargan} & 81.04 & 36.87 & 99.03 & 47.46\\
        StyleD~\cite{kim2022style} & 96.41 & 78.54 & 104.79 & 65.31\\
        \midrule
        He et al.~\cite{he2021context} & 68.97 & 24.22 & 113.58 & 54.80\\
        Pang et al.~\cite{pang2022nsd} & 71.83 & 26.64 & 99.31 & 41.28\\
        \midrule
        Ours & \bf{64.55} & \bf{11.61} & \bf{76.81} & \bf{6.30}\\
        \bottomrule
    \end{tabular}
    \caption{Quantitative results. Note that He et al.~\cite{he2021context} and Pang et al.~\cite{pang2022nsd} require explicit object layout for training and inference. Lower FID/KID scores indicate higher image generation quality.}
    \label{tab:quantitative_results}
\end{table}

\begin{figure*}[t]
    \includegraphics[width=0.24\textwidth]{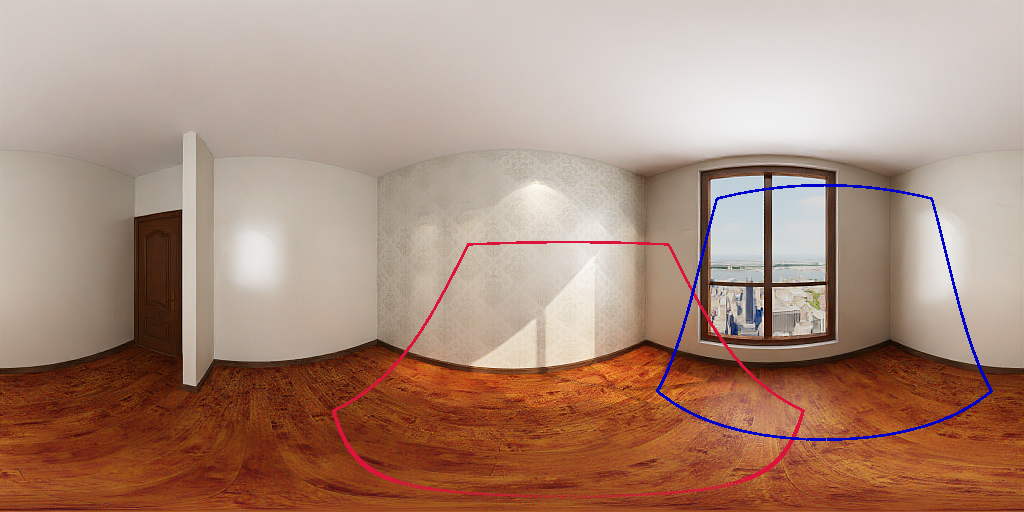}
    \hfill
    \includegraphics[width=0.24\textwidth]{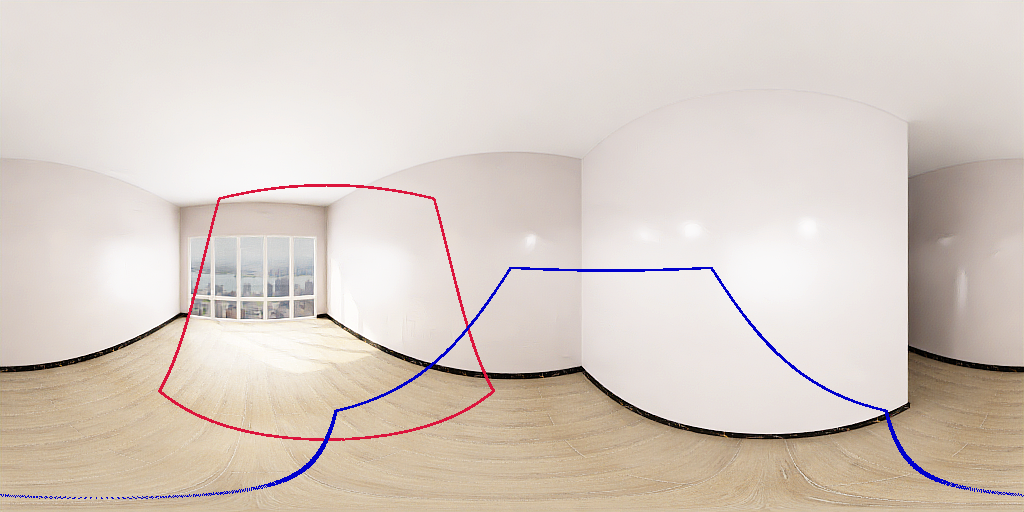}
    \hfill
    \includegraphics[width=0.24\textwidth]{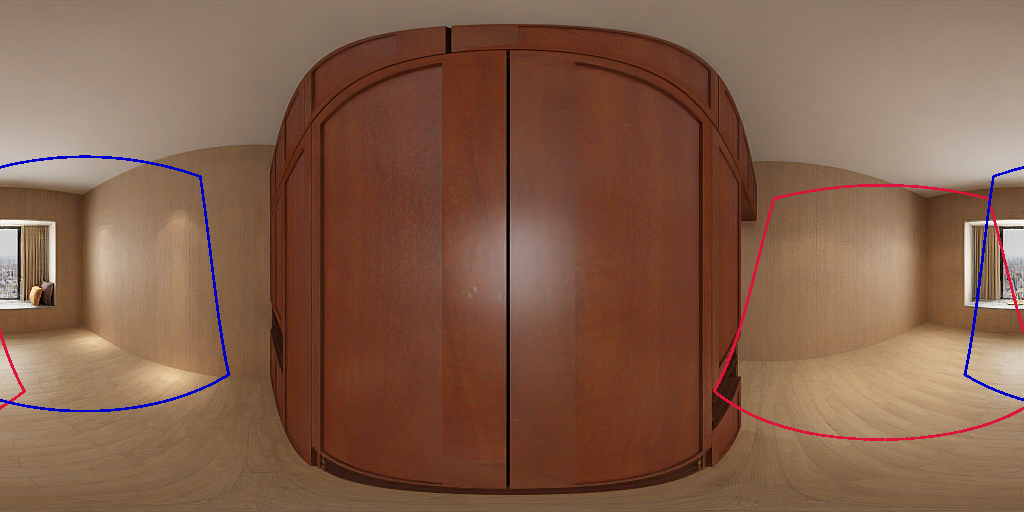}
    \hfill
    \includegraphics[width=0.24\textwidth]{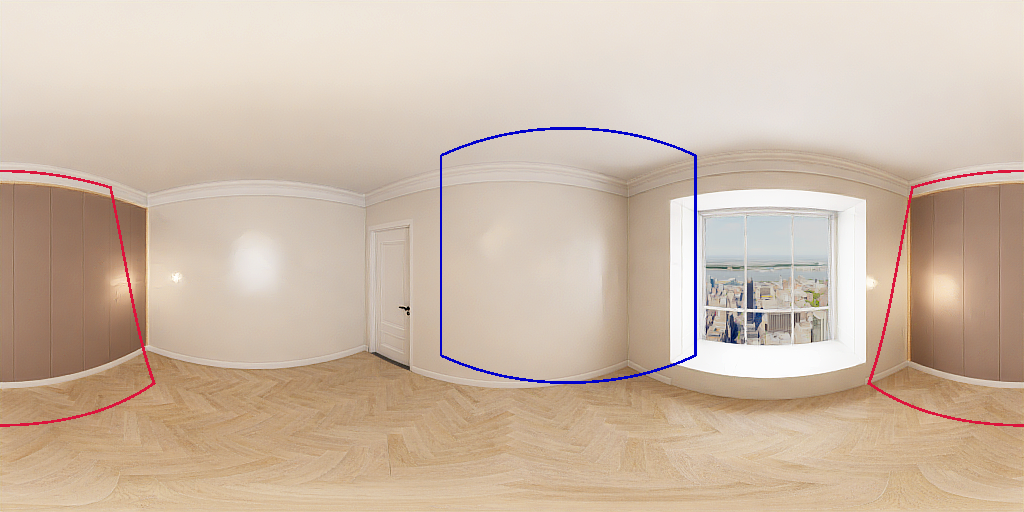}
   \\[1mm]
    \subfloat[Input]{\begin{minipage}[c]{\textwidth}
        \includegraphics[width=0.115\textwidth]{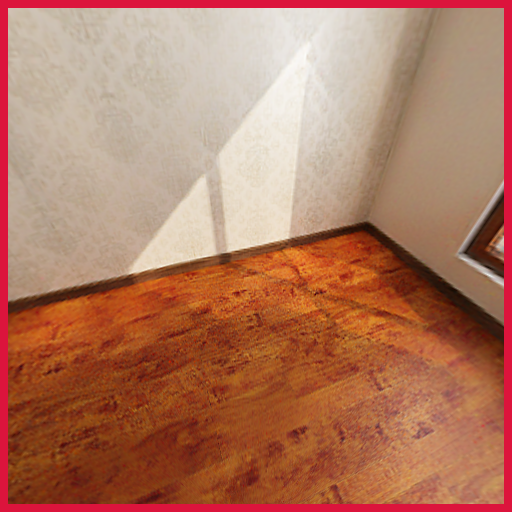}
        \hspace{0.0005\textwidth}
        \includegraphics[width=0.115\textwidth]{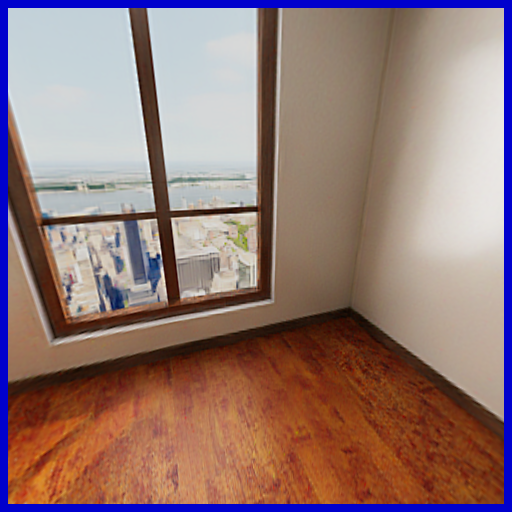}
        \hfill
        \includegraphics[width=0.115\textwidth]{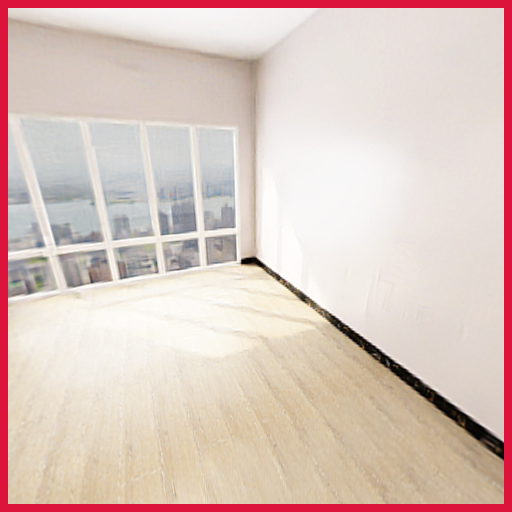}
        \hspace{0.0005\textwidth}
        \includegraphics[width=0.115\textwidth]{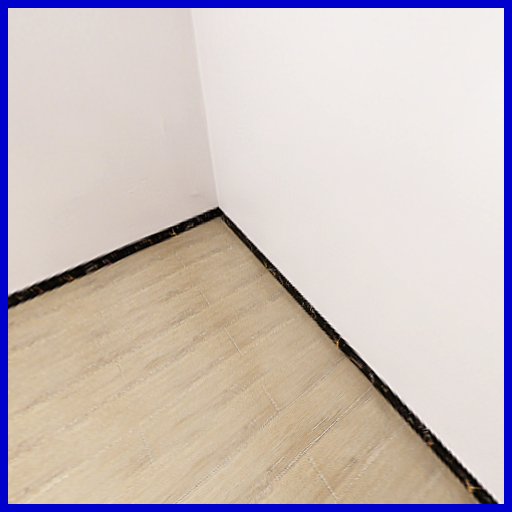}
        \hfill
        \includegraphics[width=0.115\textwidth]{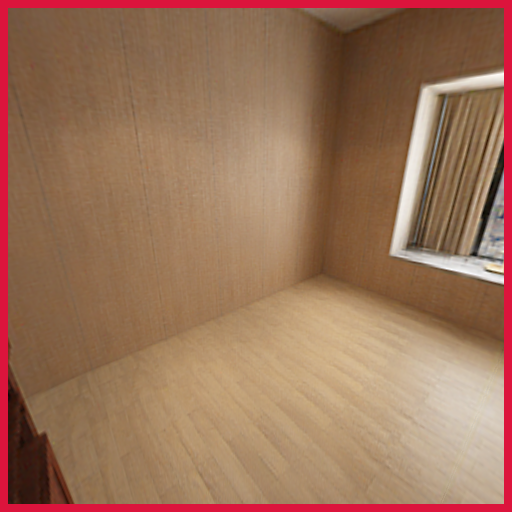}
        \hspace{0.0005\textwidth}
        \includegraphics[width=0.115\textwidth]{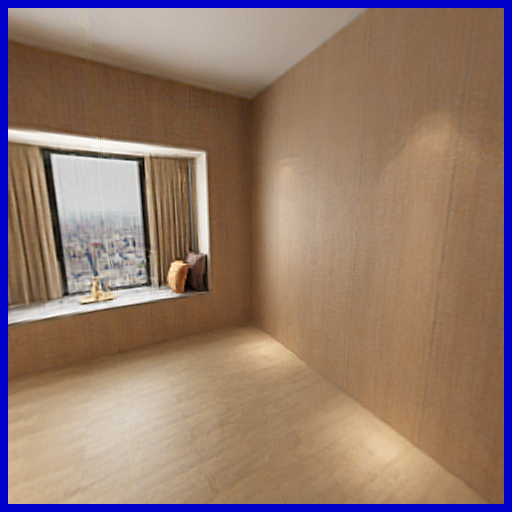}
        \hfill
        \includegraphics[width=0.115\textwidth]{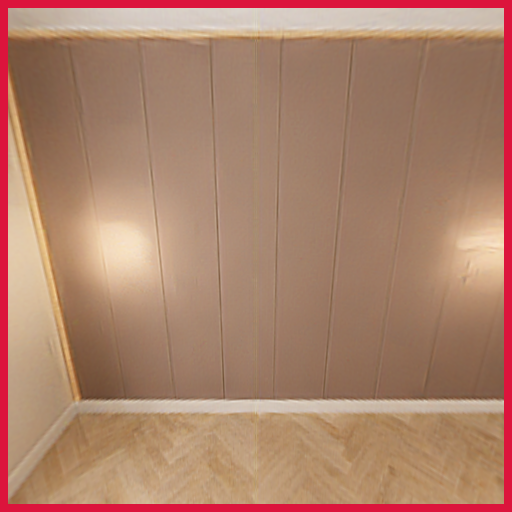}
        \hspace{0.0005\textwidth}
        \includegraphics[width=0.115\textwidth]{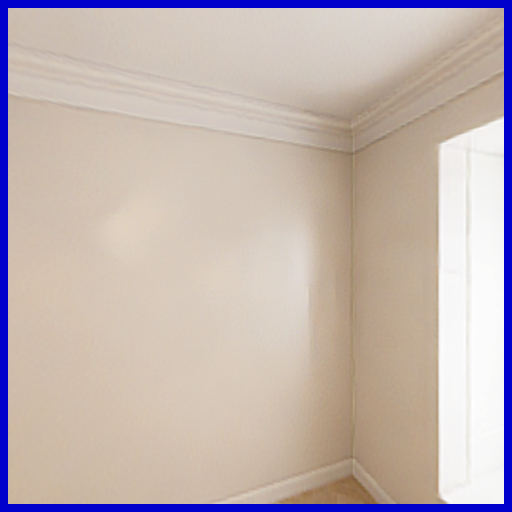}
    \end{minipage}}
    \smallskip
    
    \subfloat[Pix2PixHD]{\begin{minipage}[c]{\textwidth}
        \includegraphics[width=0.115\textwidth]{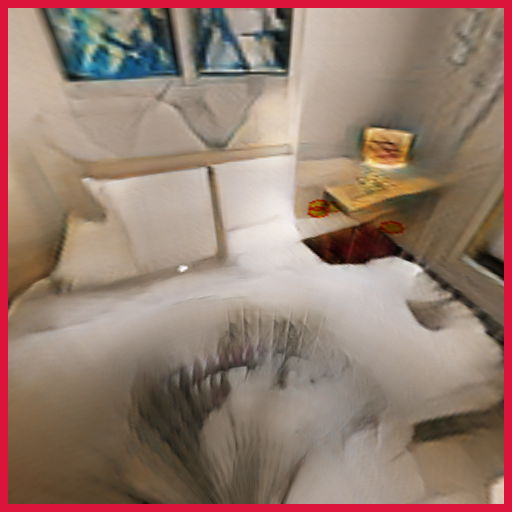}
        \hspace{0.0005\textwidth}
        \includegraphics[width=0.115\textwidth]{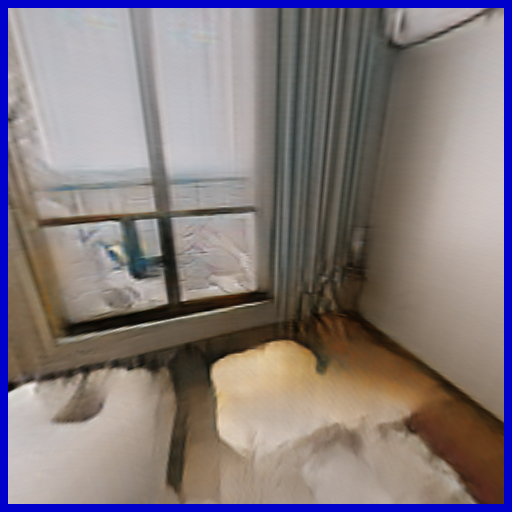}
        \hfill
        \includegraphics[width=0.115\textwidth]{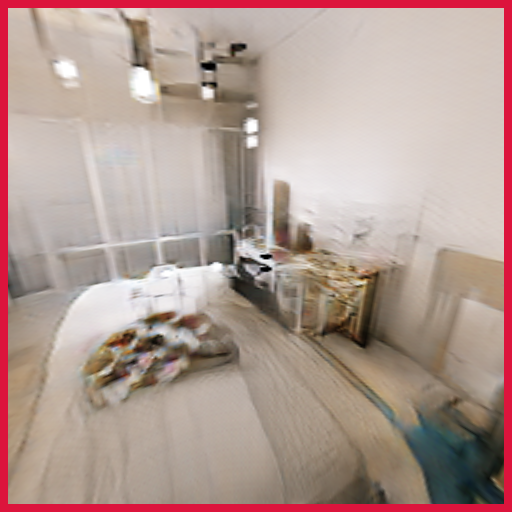}
        \hspace{0.0005\textwidth}
        \includegraphics[width=0.115\textwidth]{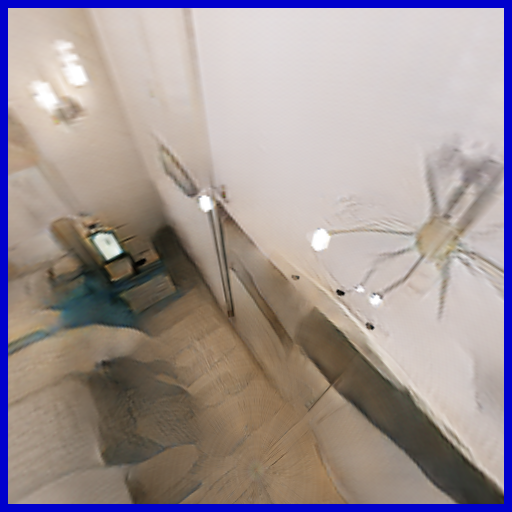}
        \hfill
        \includegraphics[width=0.115\textwidth]{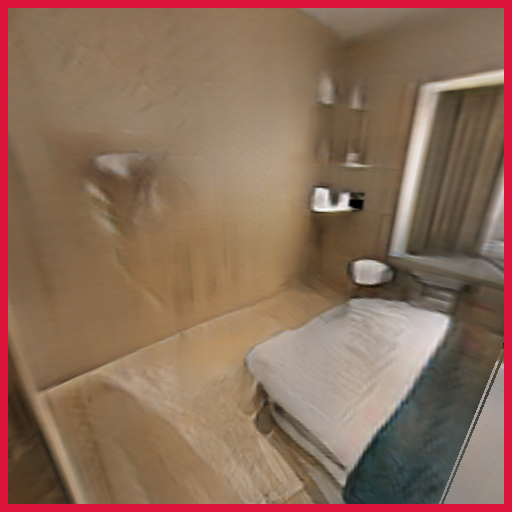}
        \hspace{0.0005\textwidth}
        \includegraphics[width=0.115\textwidth]{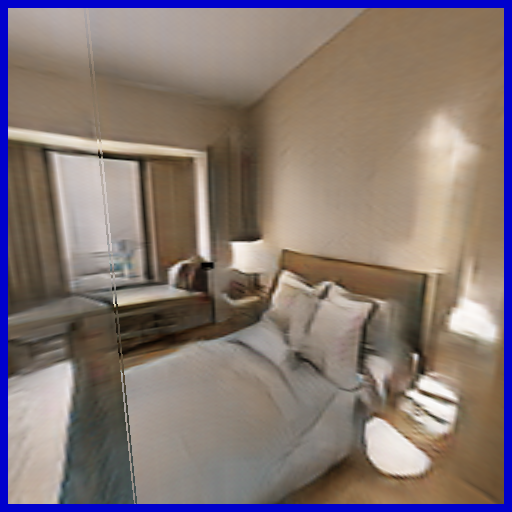}
        \hfill
        \includegraphics[width=0.115\textwidth]{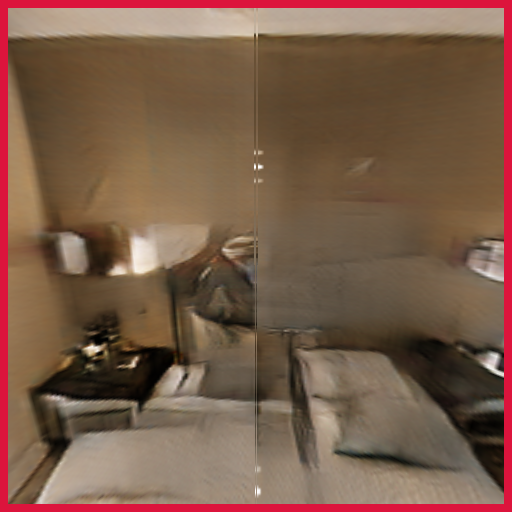}
        \hspace{0.0005\textwidth}
        \includegraphics[width=0.115\textwidth]{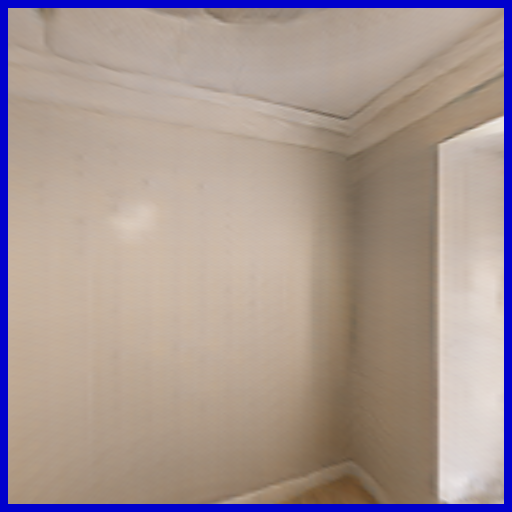}
    \end{minipage}}
    \smallskip
    
    \subfloat[StarGANv2]{\begin{minipage}[c]{\textwidth}
        \includegraphics[width=0.115\textwidth]{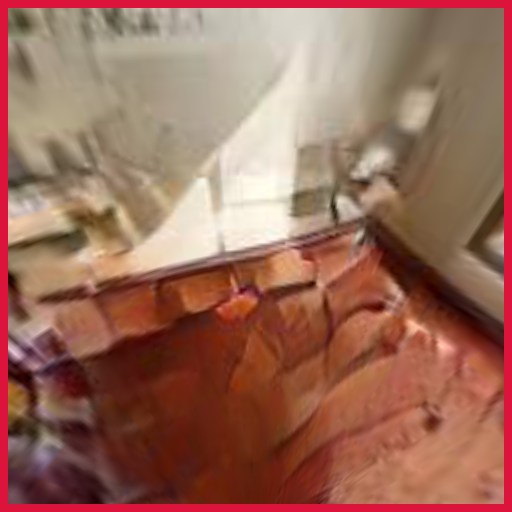}
        \hspace{0.0005\textwidth}
        \includegraphics[width=0.115\textwidth]{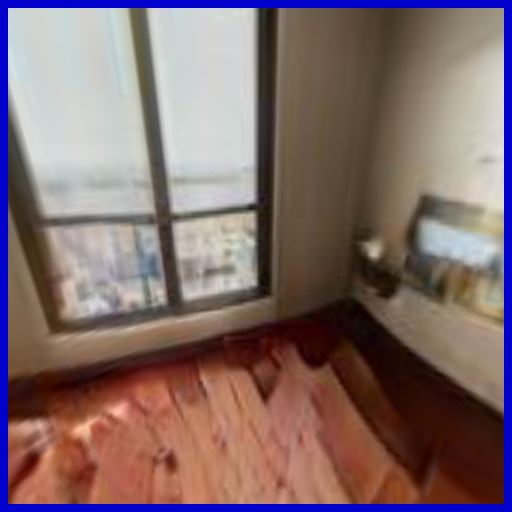}
        \hfill
        \includegraphics[width=0.115\textwidth]{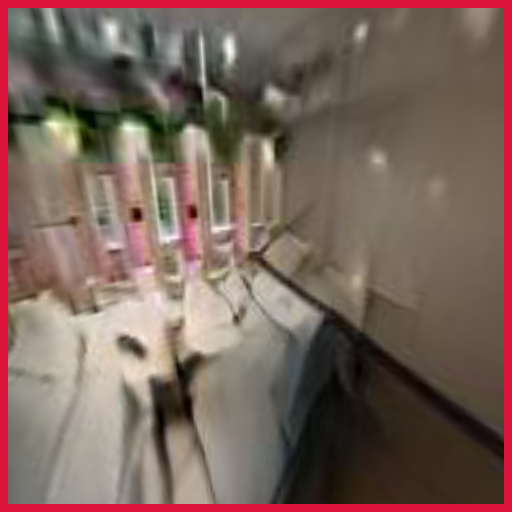}
        \hspace{0.0005\textwidth}
        \includegraphics[width=0.115\textwidth]{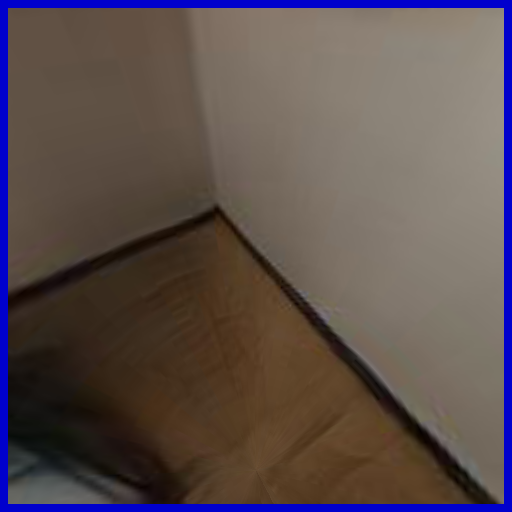}
        \hfill
        \includegraphics[width=0.115\textwidth]{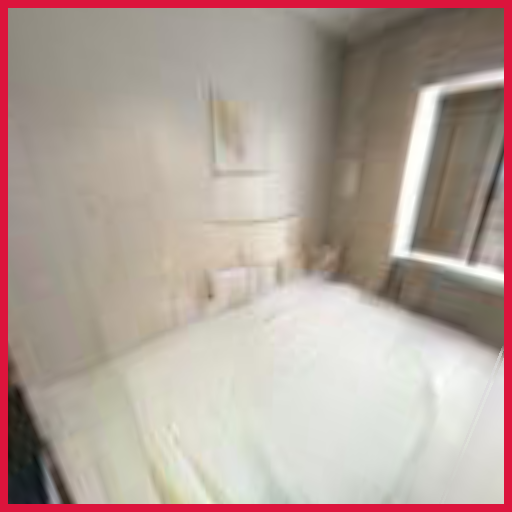}
        \hspace{0.0005\textwidth}
        \includegraphics[width=0.115\textwidth]{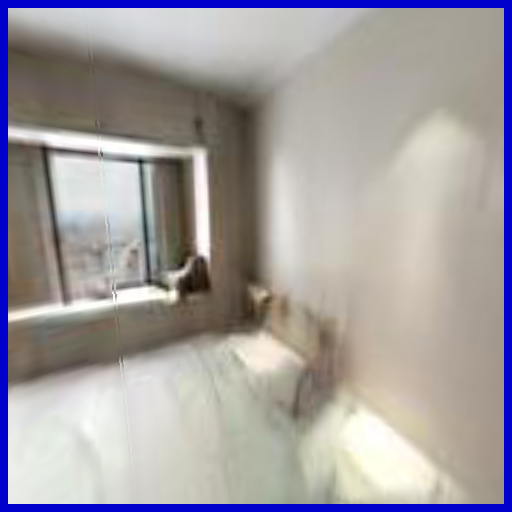}
        \hfill
        \includegraphics[width=0.115\textwidth]{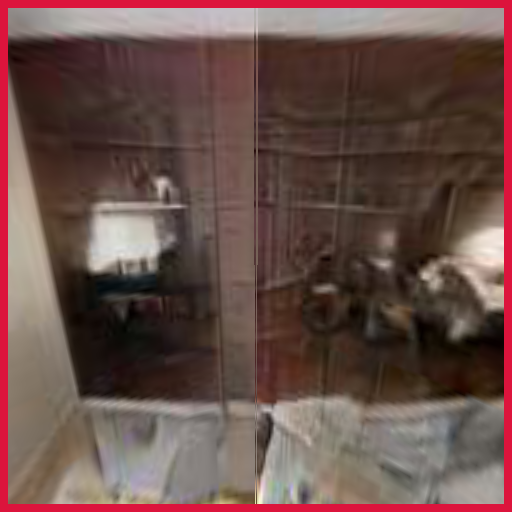}
        \hspace{0.0005\textwidth}
        \includegraphics[width=0.115\textwidth]{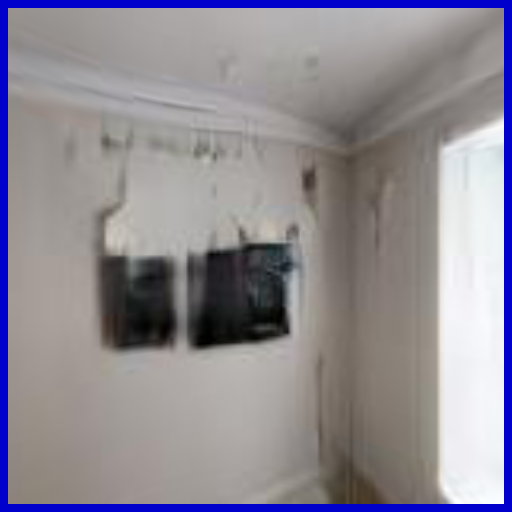}
    \end{minipage}}
    \smallskip
    
    \subfloat[StyleD]{\begin{minipage}[c]{\textwidth}
        \includegraphics[width=0.115\textwidth]{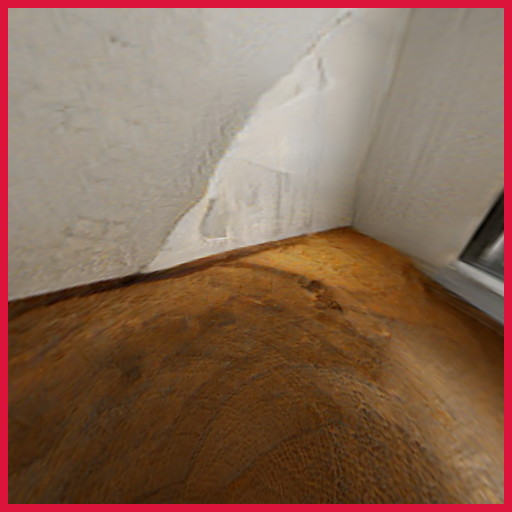}
        \hspace{0.0005\textwidth}
        \includegraphics[width=0.115\textwidth]{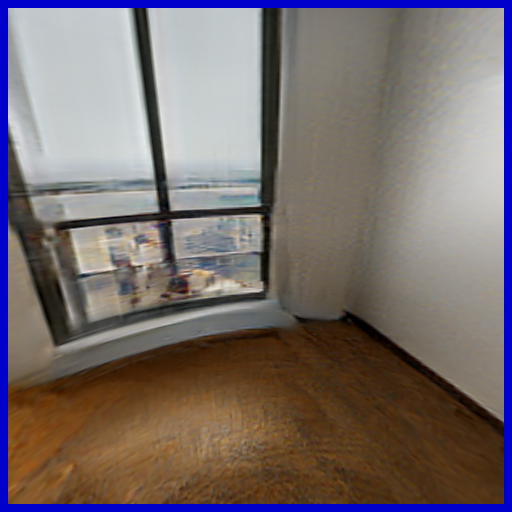}
        \hfill
        \includegraphics[width=0.115\textwidth]{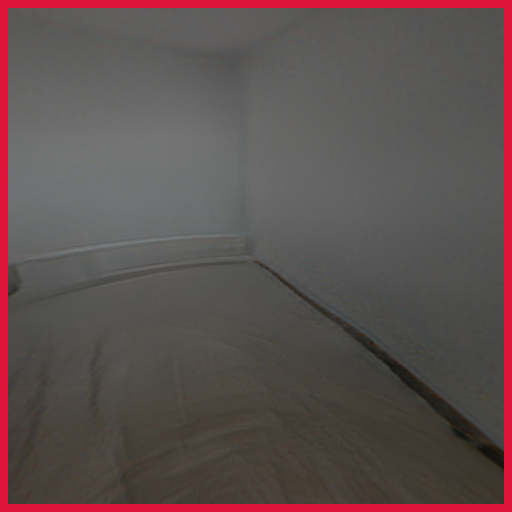}
        \hspace{0.0005\textwidth}
        \includegraphics[width=0.115\textwidth]{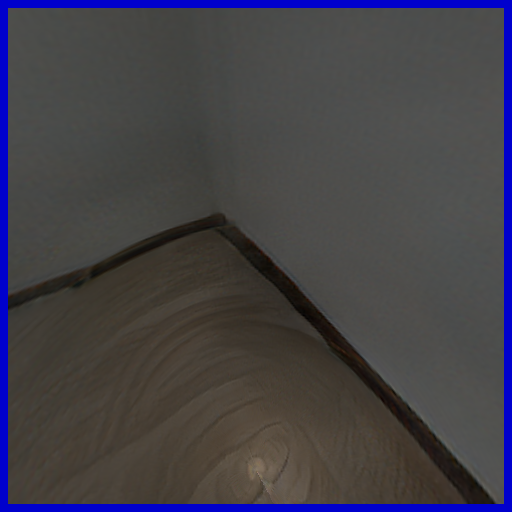}
        \hfill
        \includegraphics[width=0.115\textwidth]{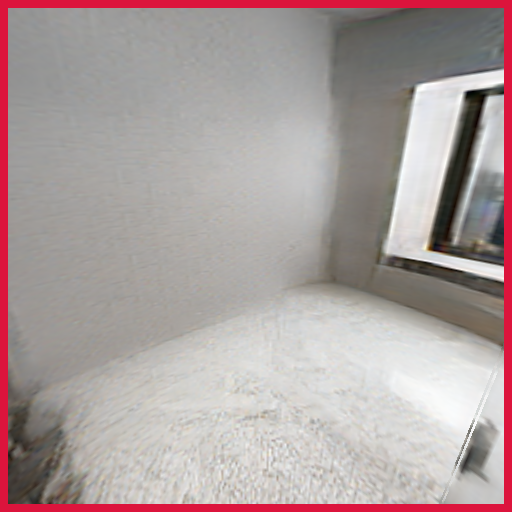}
        \hspace{0.0005\textwidth}
        \includegraphics[width=0.115\textwidth]{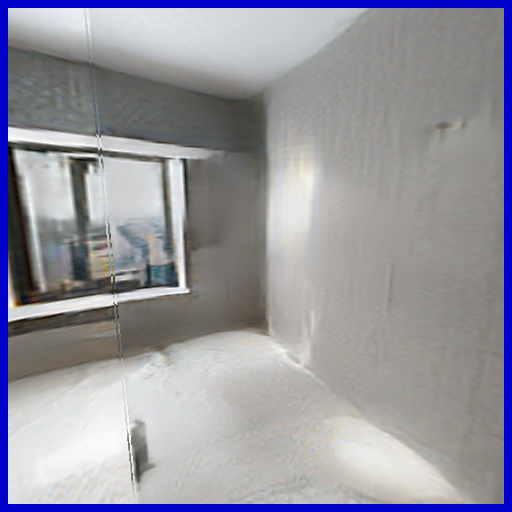}
        \hfill
        \includegraphics[width=0.115\textwidth]{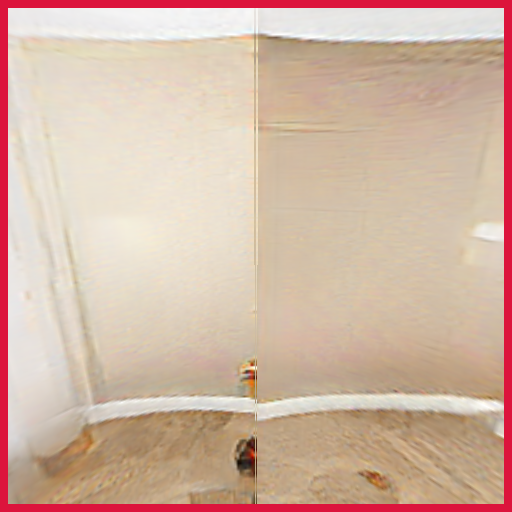}
        \hspace{0.0005\textwidth}
        \includegraphics[width=0.115\textwidth]{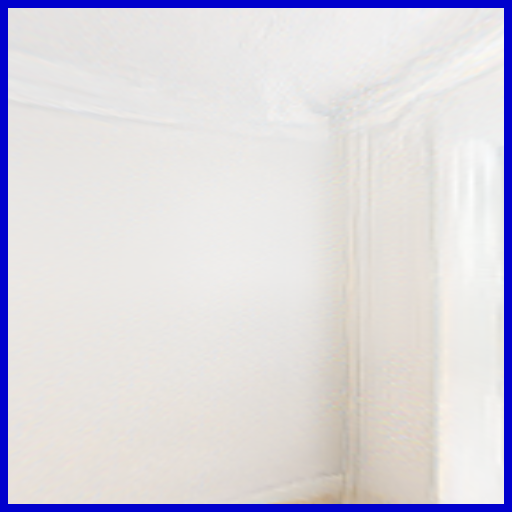}
    \end{minipage}}
    \smallskip

    \subfloat[Pang et al.~\cite{pang2022nsd} ~\textit{(with object layout)}]{\begin{minipage}[c]{\textwidth}
        \includegraphics[width=0.115\textwidth]{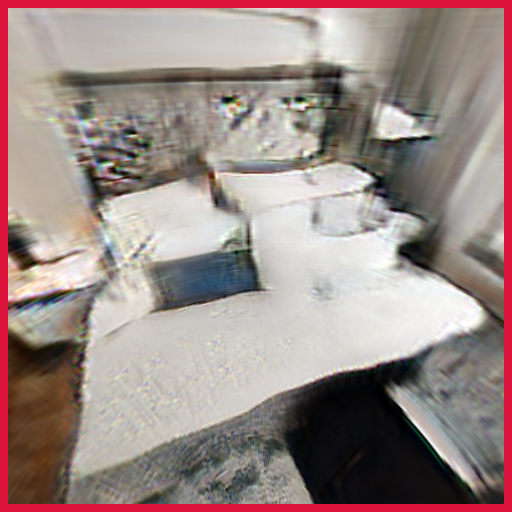}
        \hspace{0.0005\textwidth}
        \includegraphics[width=0.115\textwidth]{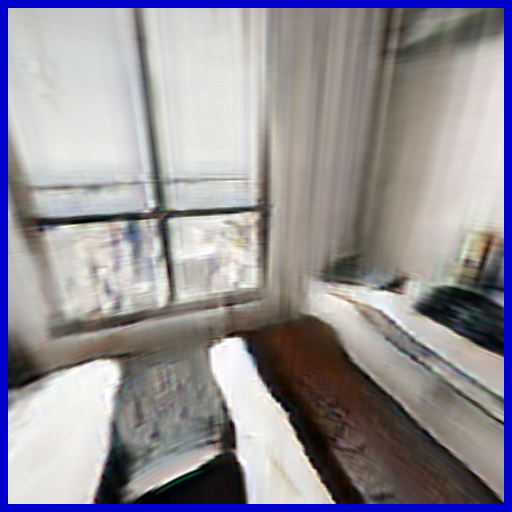}
        \hfill
        \includegraphics[width=0.115\textwidth]{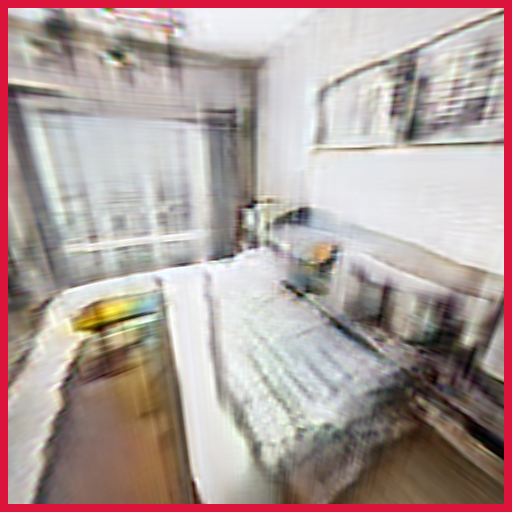}
        \hspace{0.0005\textwidth}
        \includegraphics[width=0.115\textwidth]{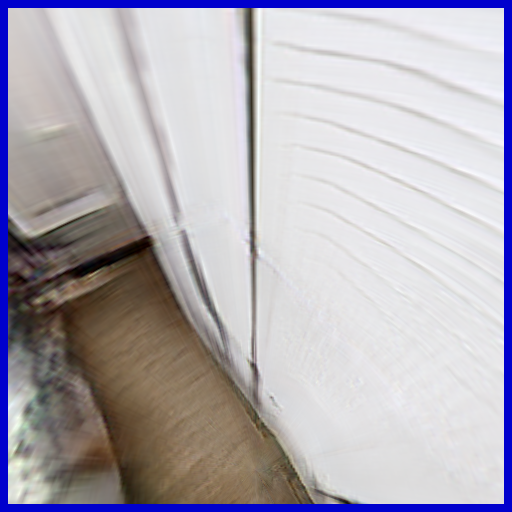}
        \hfill
        \includegraphics[width=0.115\textwidth]{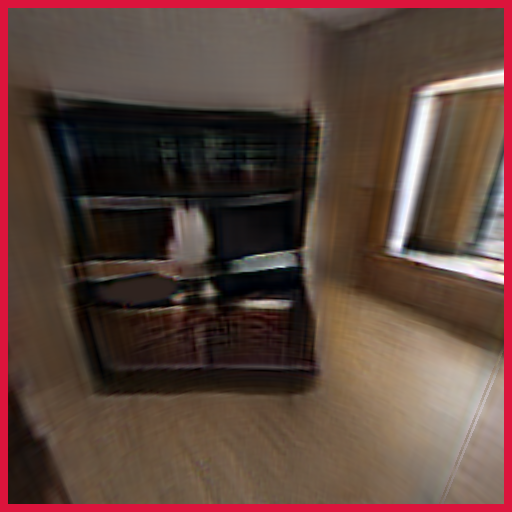}
        \hspace{0.0005\textwidth}
        \includegraphics[width=0.115\textwidth]{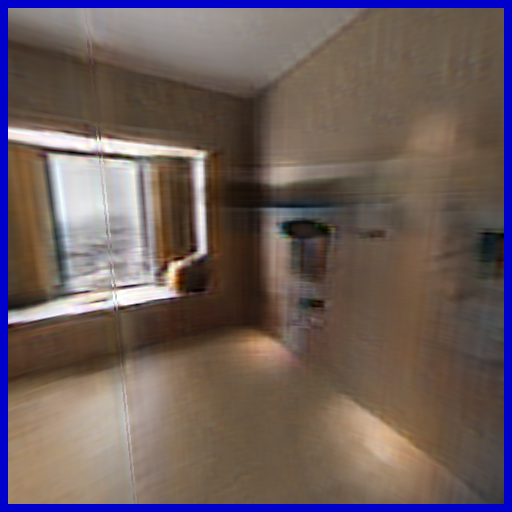}
        \hfill
        \includegraphics[width=0.115\textwidth]{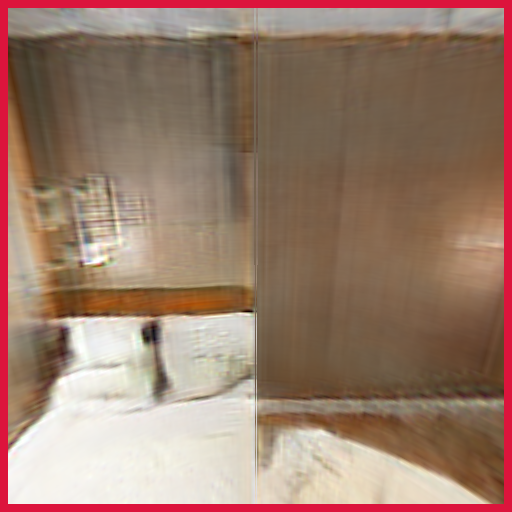}
        \hspace{0.0005\textwidth}
        \includegraphics[width=0.115\textwidth]{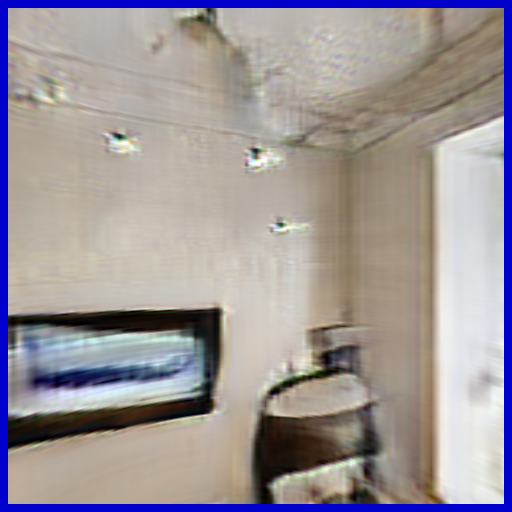}
    \end{minipage}}
    \smallskip

    \subfloat[He et al.~\cite{he2021context} ~\textit{(with object layout)}]{\begin{minipage}[c]{\textwidth}
        \includegraphics[width=0.115\textwidth]{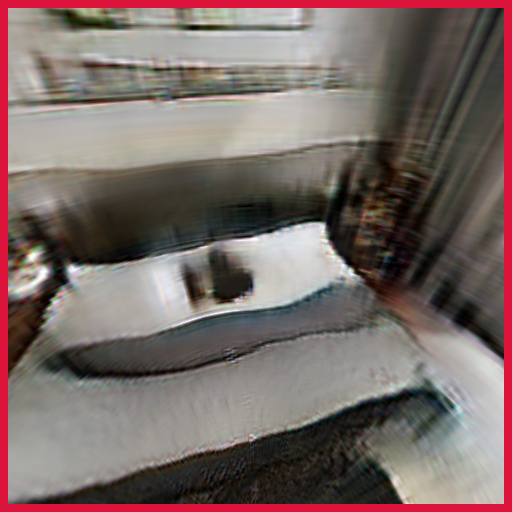}
        \hspace{0.0005\textwidth}
        \includegraphics[width=0.115\textwidth]{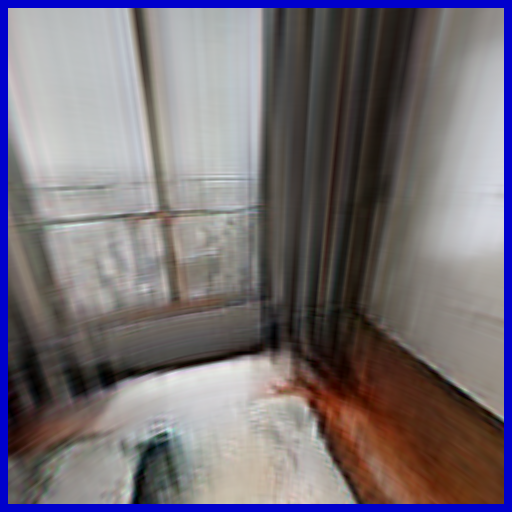}
        \hfill
        \includegraphics[width=0.115\textwidth]{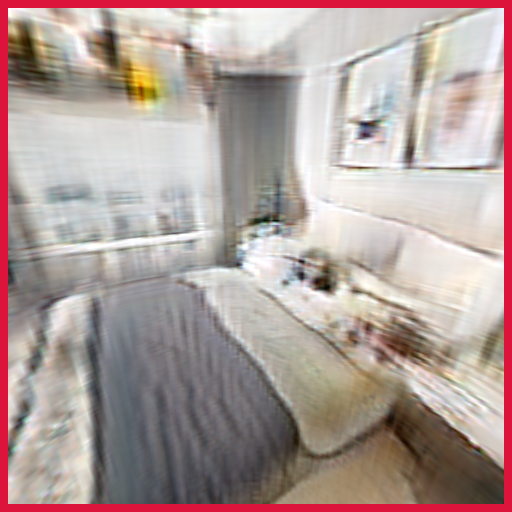}
        \hspace{0.0005\textwidth}
        \includegraphics[width=0.115\textwidth]{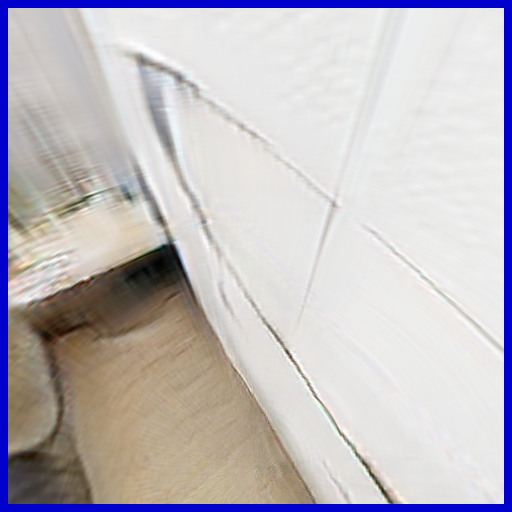}
        \hfill
        \includegraphics[width=0.115\textwidth]{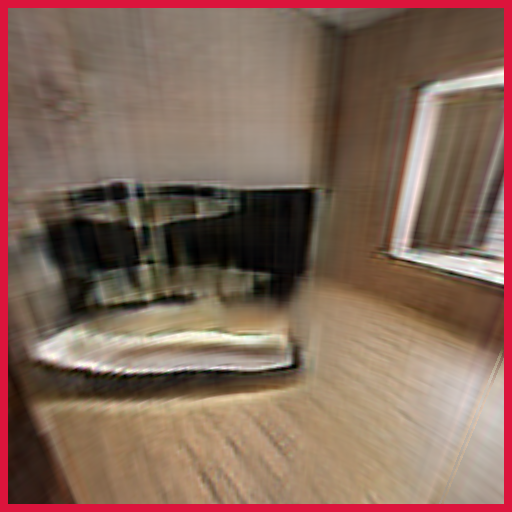}
        \hspace{0.0005\textwidth}
        \includegraphics[width=0.115\textwidth]{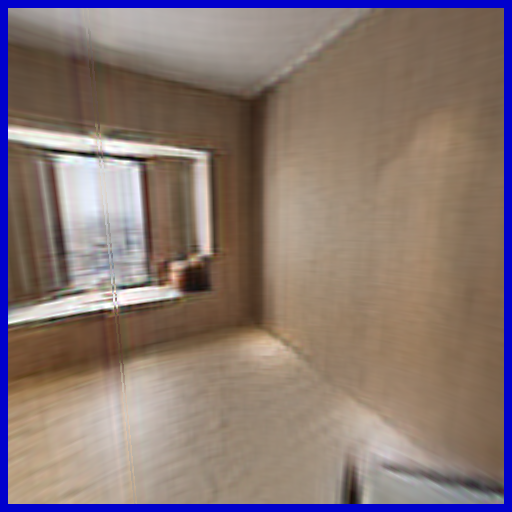}
        \hfill
        \includegraphics[width=0.115\textwidth]{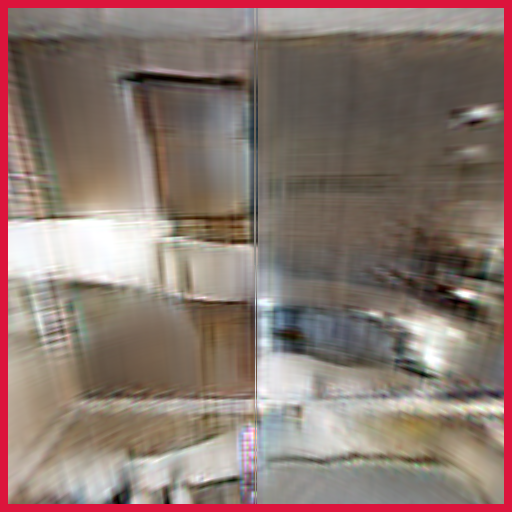}
        \hspace{0.0005\textwidth}
        \includegraphics[width=0.115\textwidth]{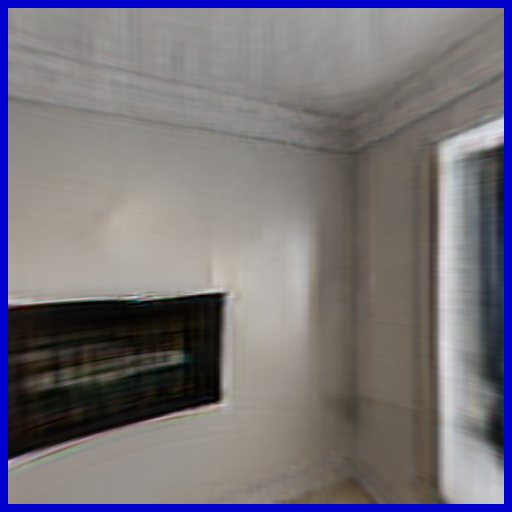}
    \end{minipage}}
    \smallskip

    \subfloat[Ours]{\begin{minipage}[c]{\textwidth}
        \includegraphics[width=0.115\textwidth]{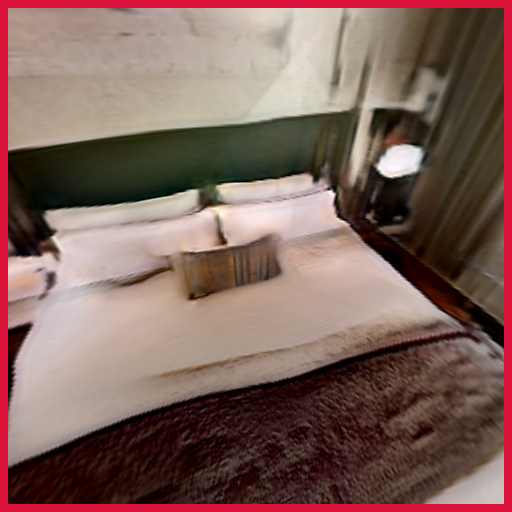}
        \hspace{0.0005\textwidth}
        \includegraphics[width=0.115\textwidth]{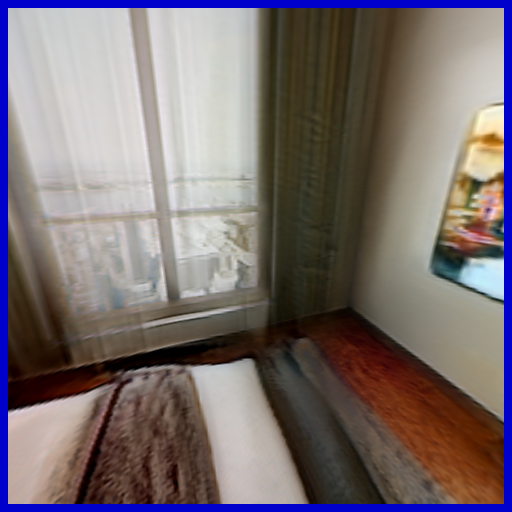}
        \hfill
        \includegraphics[width=0.115\textwidth]{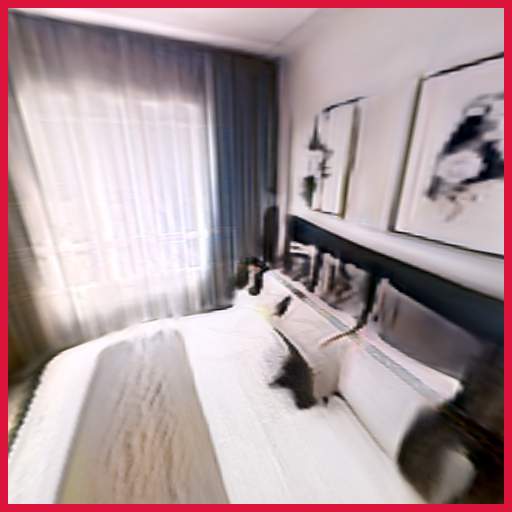}
        \hspace{0.0005\textwidth}
        \includegraphics[width=0.115\textwidth]{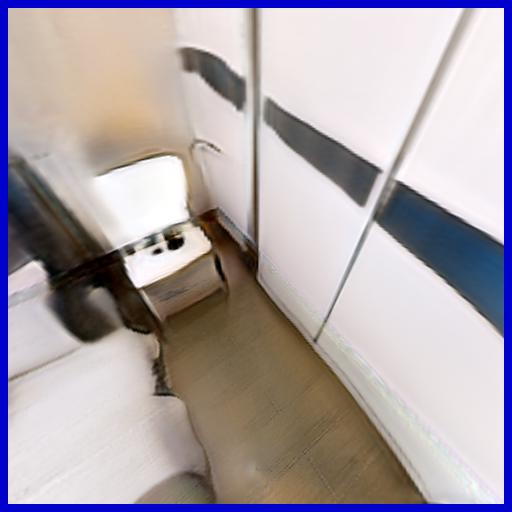}
        \hfill
        \includegraphics[width=0.115\textwidth]{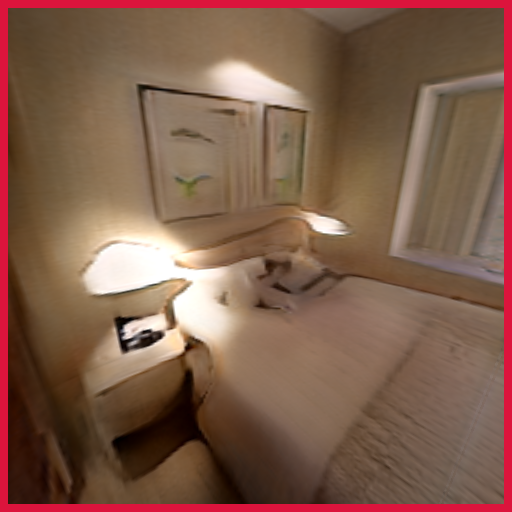}
        \hspace{0.0005\textwidth}
        \includegraphics[width=0.115\textwidth]{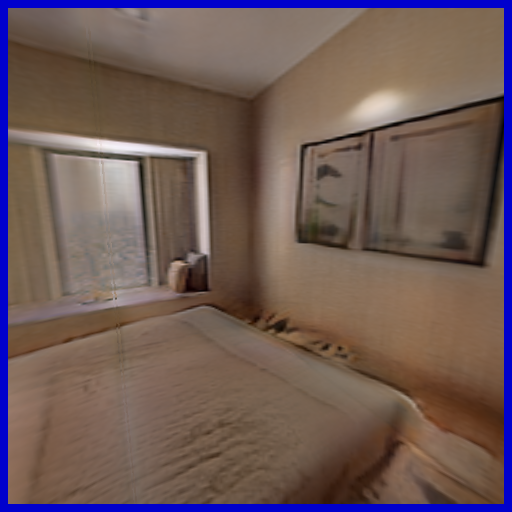}
        \hfill
        \includegraphics[width=0.115\textwidth]{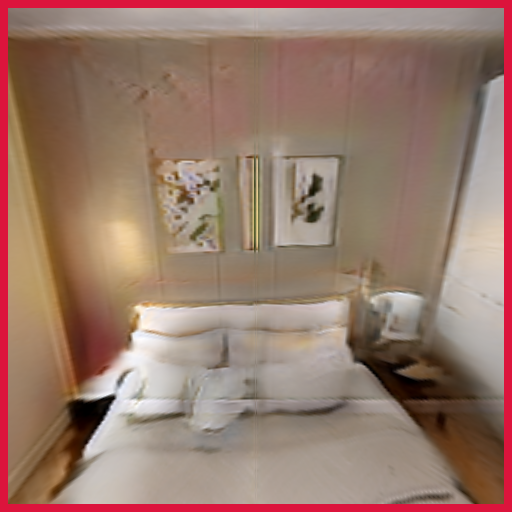}
        \hspace{0.0005\textwidth}
        \includegraphics[width=0.115\textwidth]{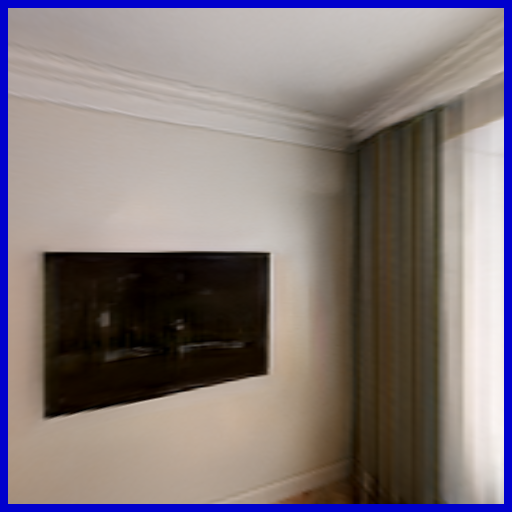}
    \end{minipage}}
    \smallskip
    
    \caption{Visualization of the generated 360\degree images. Compared to ours, Pang et al.~\cite{pang2022nsd} and He et al.~\cite{he2021context} require an additional explicit object layout as input. More results are in the supplementary material.} 
    \label{fig:qualitative_comparison}
\end{figure*}

\noindent\textbf{Qualitative results.}
We qualitatively compare our method with the baselines in Figure~\ref{fig:qualitative_comparison}. As shown in the results, our method generates photo-realistic images in the 360\degree viewer with plausible furniture arrangements. Background details in input images are well maintained. More importantly, while I2I baselines show difficulty in generating objects in the 360\degree setting, our model can create decent results with proper object distribution without using any explicit object labels. Compared with other layout-based generation results, our results also have fewer visual artifacts and more realistic object texture.

\begin{figure}
  \begin{subfigure}{0.49\columnwidth}
    \includegraphics[width=\textwidth]{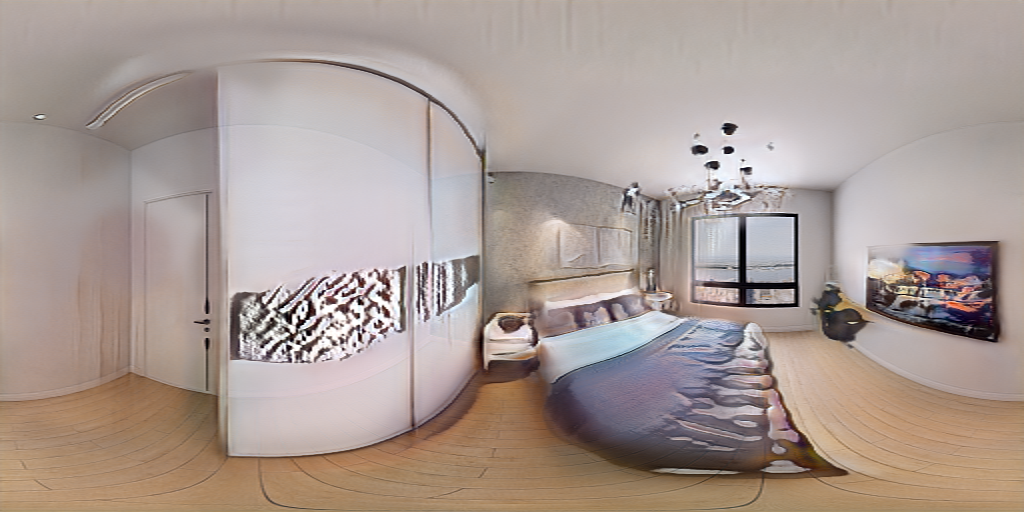}
  \end{subfigure}
  \hfill 
  \begin{subfigure}{0.49\columnwidth}
    \includegraphics[width=\textwidth]{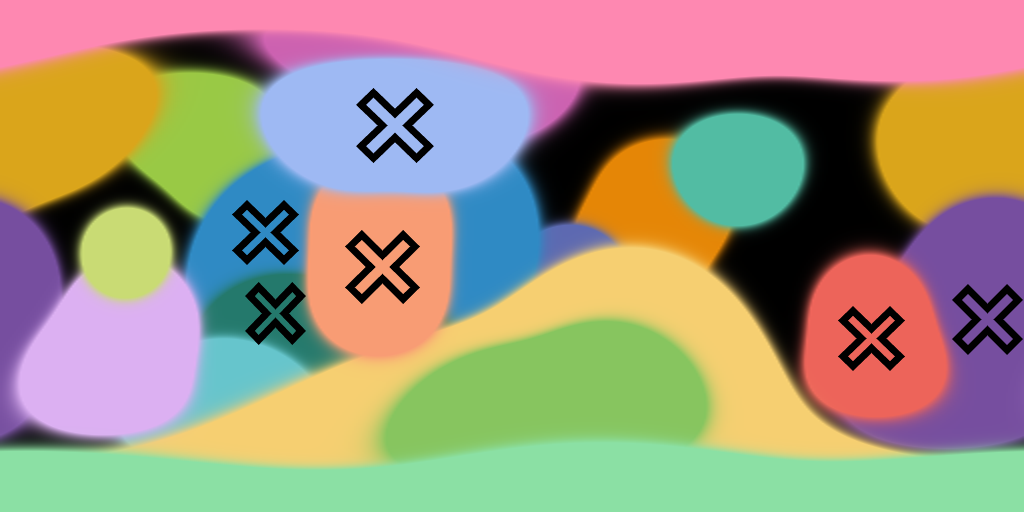}
  \end{subfigure}
  \\[1mm]
  \subfloat[Remove wardrobe and remove TV]{\begin{minipage}[c]{\columnwidth}
      \begin{subfigure}{0.49\columnwidth}
        \includegraphics[width=\textwidth]{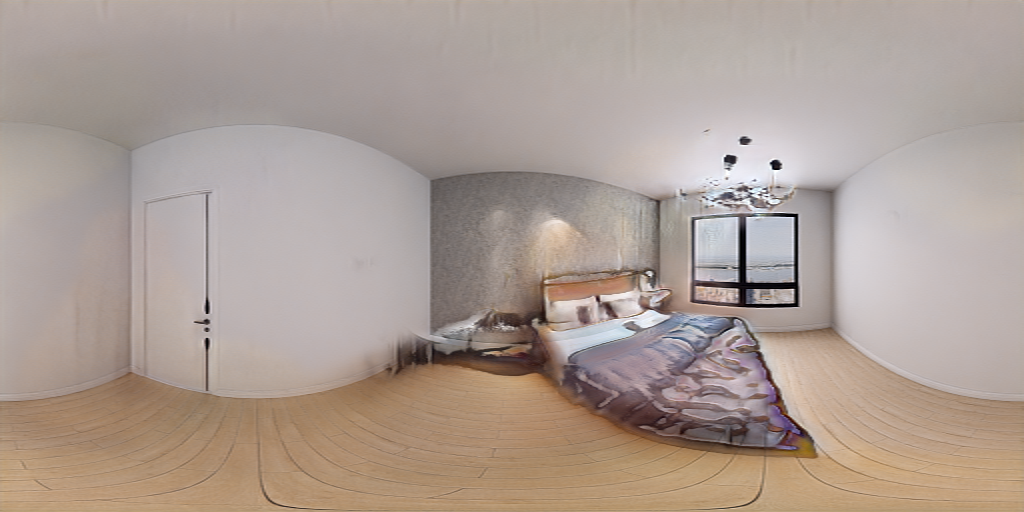}
      \end{subfigure}
      \hfill 
      \begin{subfigure}{0.49\columnwidth}
        \includegraphics[width=\textwidth]{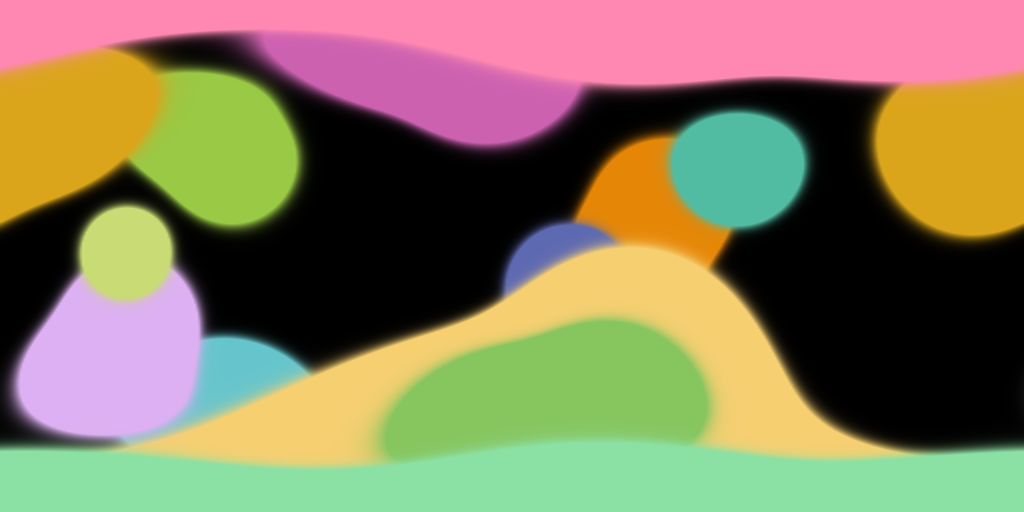}
      \end{subfigure}
  \end{minipage}}
  \\[2mm]
  \begin{subfigure}{0.49\columnwidth}
    \includegraphics[width=\textwidth]{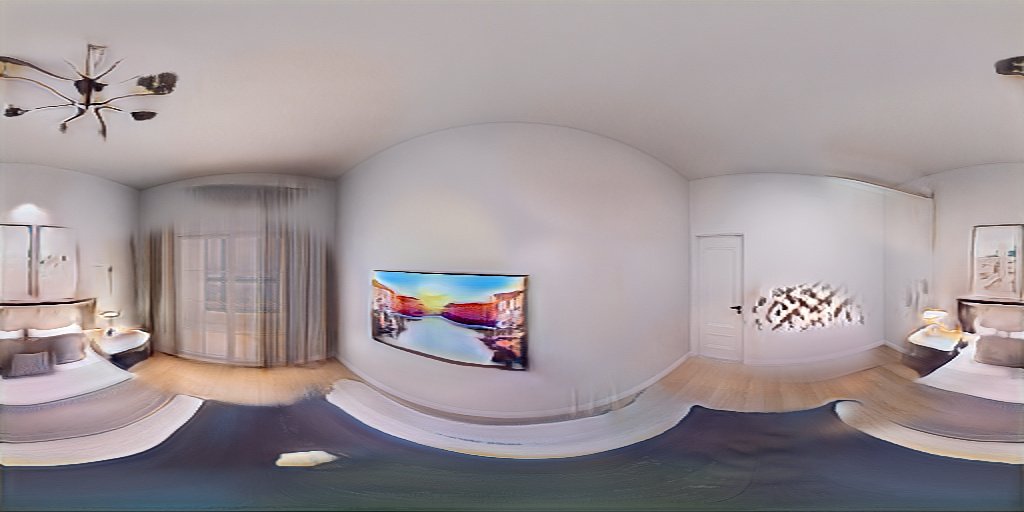}
  \end{subfigure}
  \hfill 
  \begin{subfigure}{0.49\columnwidth}
    \includegraphics[width=\textwidth]{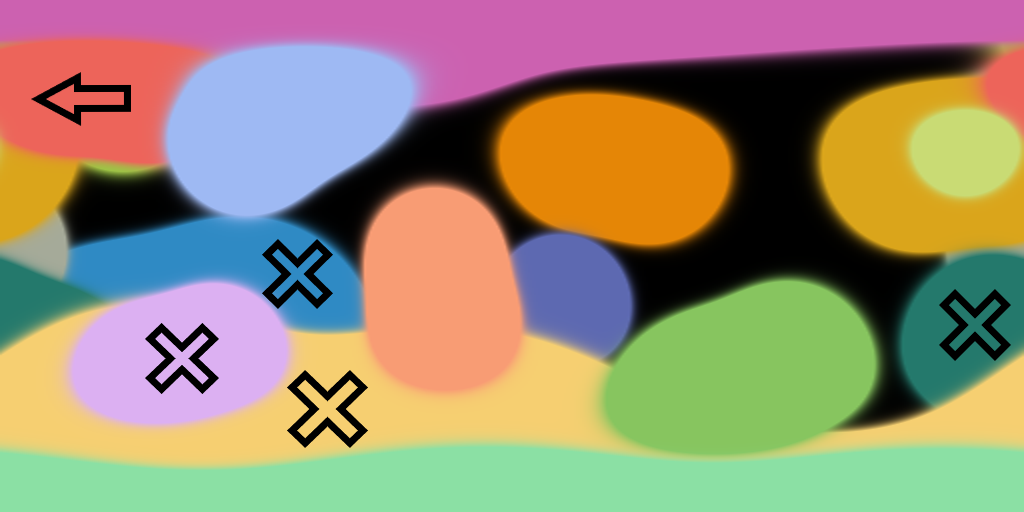}
  \end{subfigure}
  \\[1mm]
  \subfloat[Shift lamp layout left (move to the camera top right) and remove bed]{\begin{minipage}[c]{\columnwidth}
      \begin{subfigure}{0.49\columnwidth}
        \includegraphics[width=\textwidth]{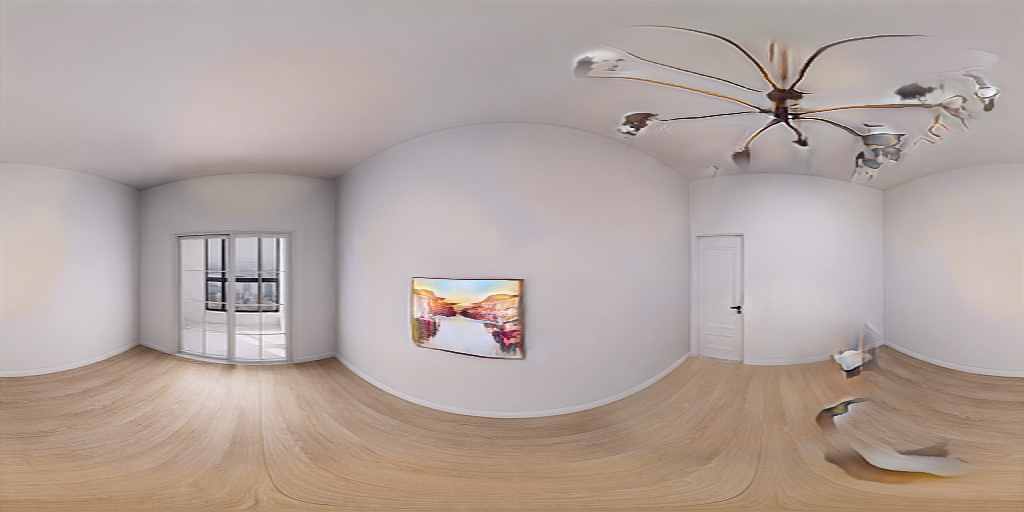}
      \end{subfigure}
      \hfill 
      \begin{subfigure}{0.49\columnwidth}
        \includegraphics[width=\textwidth]{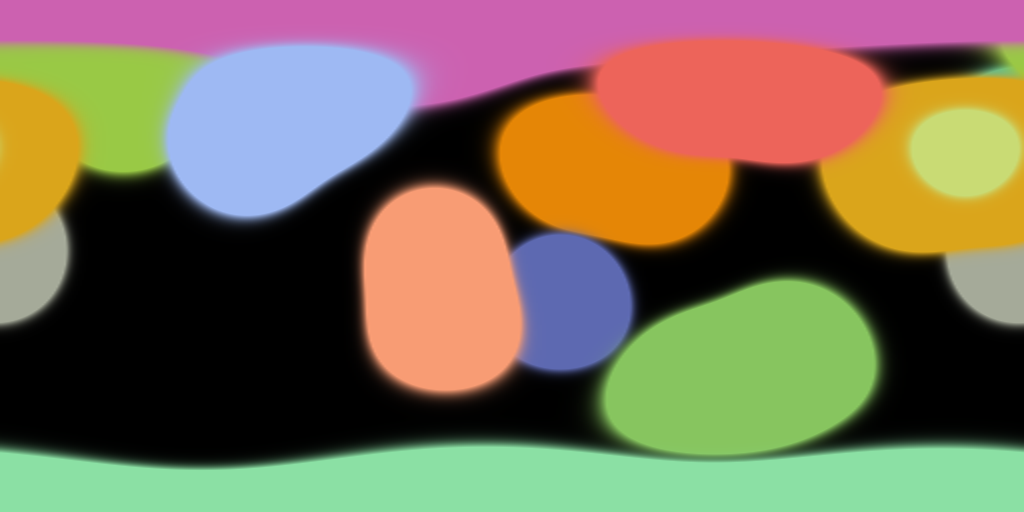}
      \end{subfigure}
  \end{minipage}}
  \\[2mm]
  \begin{subfigure}{0.49\columnwidth}
    \includegraphics[width=\textwidth]{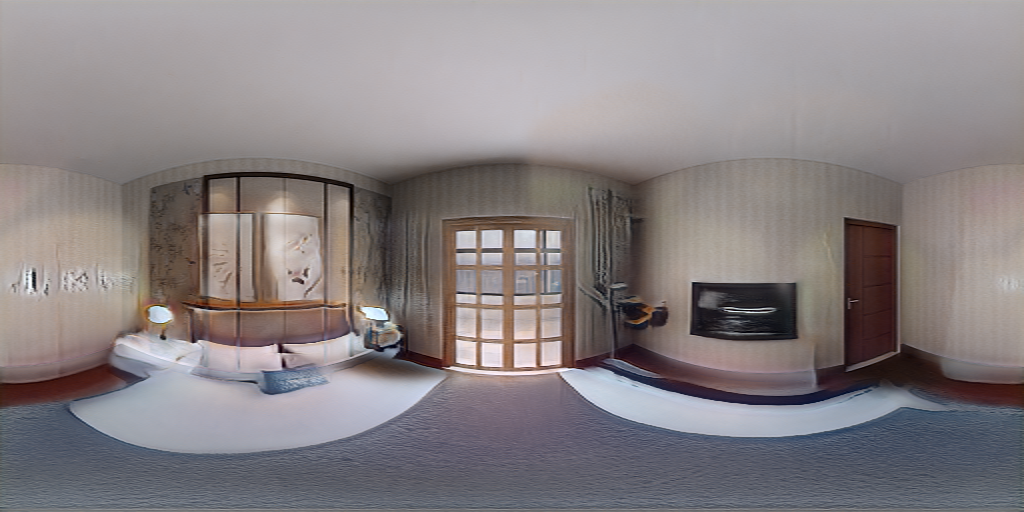}
  \end{subfigure}
  \hfill 
  \begin{subfigure}{0.49\columnwidth}
    \includegraphics[width=\textwidth]{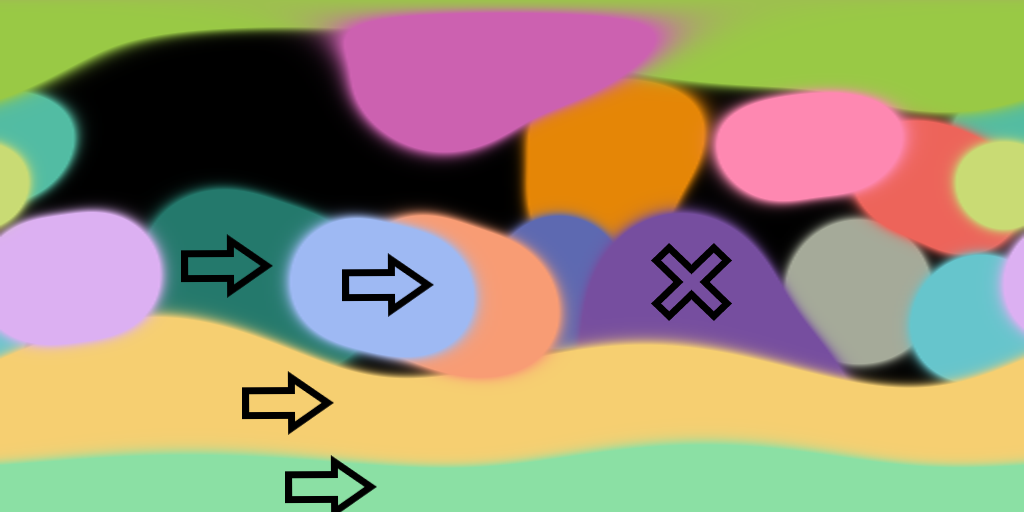}
  \end{subfigure}
  \\[1mm]
  \subfloat[Remove TV and shift bed rightward (new bed at the opposite wall)]{\begin{minipage}[c]{\columnwidth}
      \begin{subfigure}{0.49\columnwidth}
        \includegraphics[width=\textwidth]{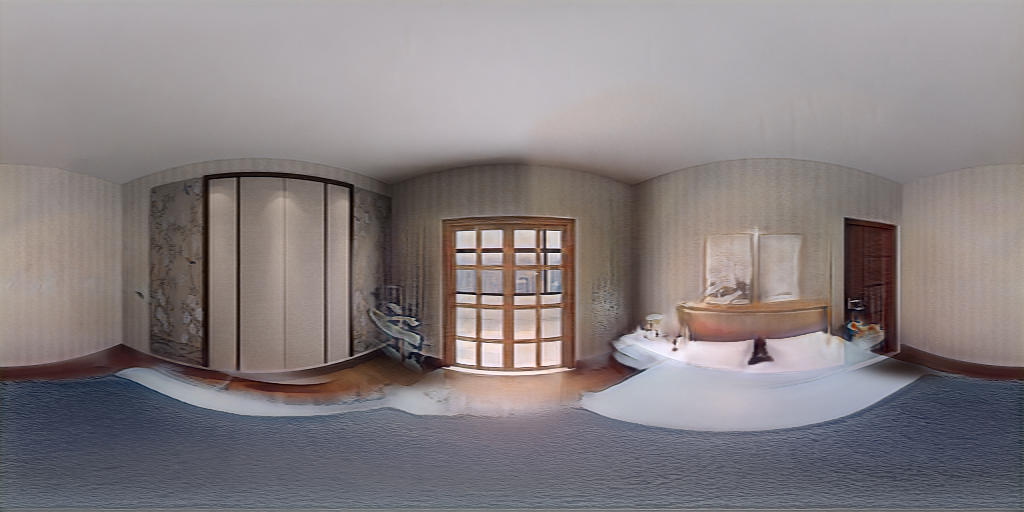}
      \end{subfigure}
      \hfill 
      \begin{subfigure}{0.49\columnwidth}
        \includegraphics[width=\textwidth]{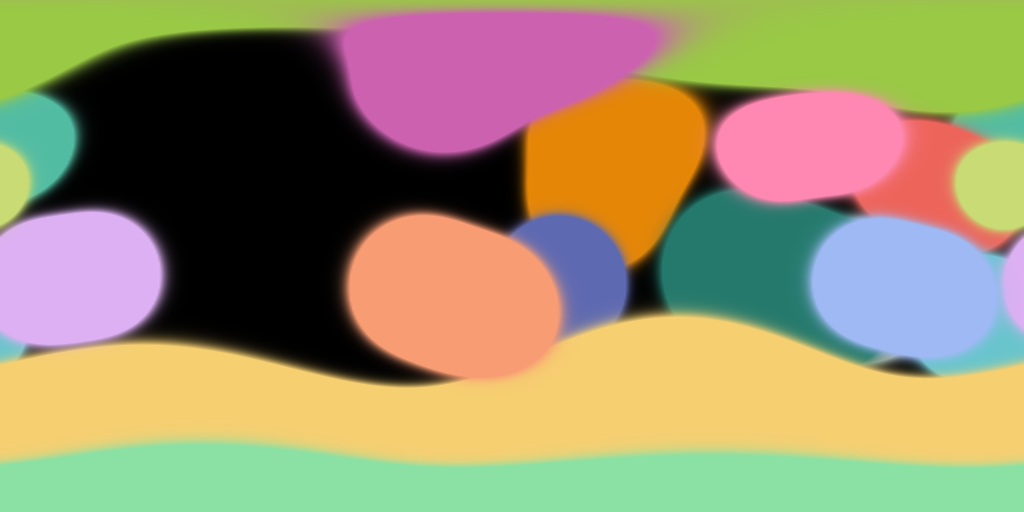}
      \end{subfigure}
  \end{minipage}}
  
\caption{Perform removal and translation manipulation on the object layout to control the generation of objects. (a), (b), and (c) show different sets of controls over different generated images. For each set of controls, the top left is the original image before manipulation, the top right is its object layout and the type of manipulation on specific object ellipses, bottom left and right are the generated image and object layout after the manipulation.}
\label{fig:controllability}
\end{figure}

\noindent\textbf{Controllability.} To illustrate the controllability of our method over the generated content, we manipulate object vectors generated by the layout generator. In particular, as our layout generator is trained in an unsupervised manner, we can only obtain the semantics of object vectors in a synthesized image \emph{after} the image is generated. We then select object ellipses in the layout for object manipulation. We observe from our results that, operations such as minimizing the object ellipse size $s$ or moving the ellipse location $(\alpha, \beta)$ result in the removal or translation of corresponding objects. Since the training of the model is conducted without explicit object labels, multiple object ellipses may contribute to a single object of a bigger size. Note that some object ellipses may not be strictly bound to any generated objects. We hypothesize that the generated object layout recommends possible furniture arrangements for the decorator to consider. The decorator may ignore some arrangements to produce a more plausible output. We illustrate the controllability of our method in Figure~\ref{fig:controllability}, which shows the diversity of generated images by manipulating the learned object layout. 


\begin{figure}
  \begin{subfigure}{0.49\columnwidth}
    \includegraphics[width=\textwidth]{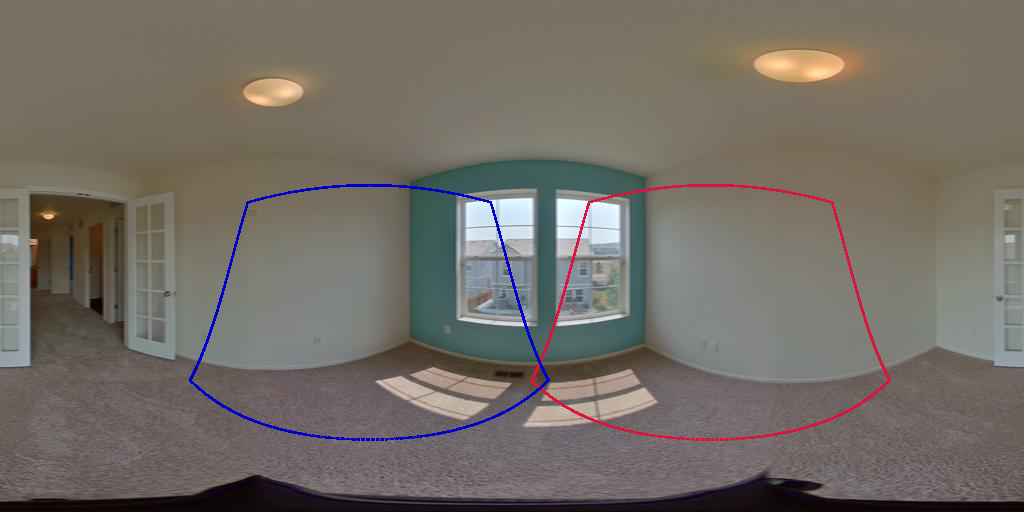}
  \end{subfigure}
  \hfill 
  \begin{subfigure}{0.49\columnwidth}
    \includegraphics[width=\textwidth]{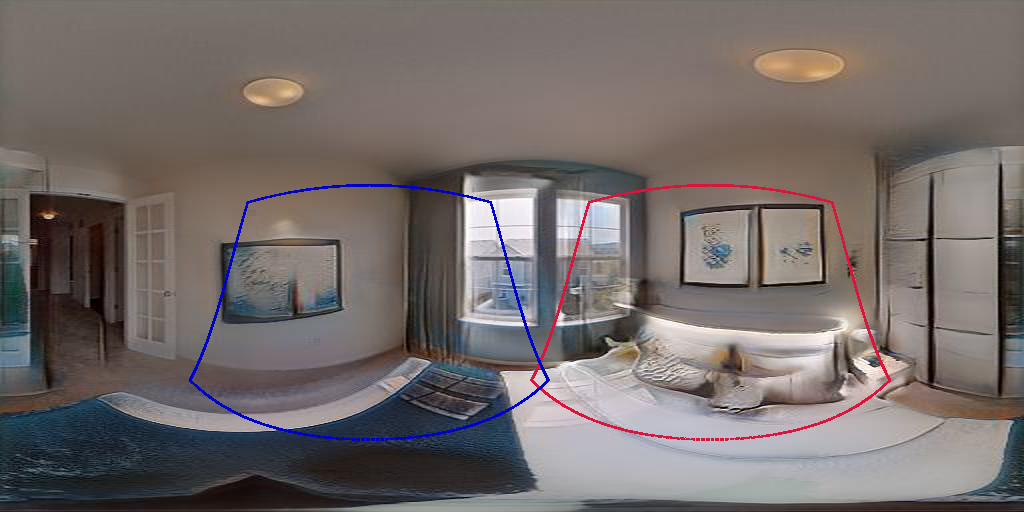}
  \end{subfigure}
  \\[0.8mm]
  \begin{minipage}[c]{0.49\columnwidth}
  \begin{subfigure}{0.48\columnwidth}
    \includegraphics[width=\textwidth]{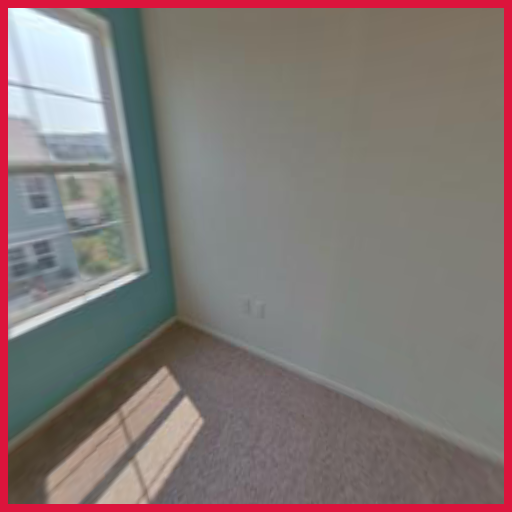}
  \end{subfigure}
  \hfill 
  \begin{subfigure}{0.48\columnwidth}
    \includegraphics[width=\textwidth]{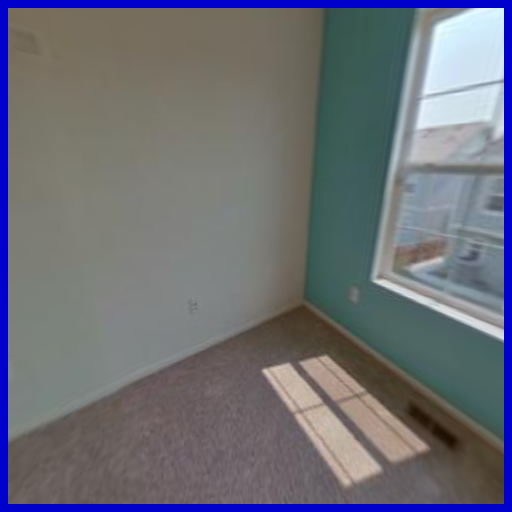}
  \end{subfigure}
  \end{minipage}
  \hfill 
  \begin{minipage}[c]{0.49\columnwidth}
  \begin{subfigure}{0.48\columnwidth}
    \includegraphics[width=\textwidth]{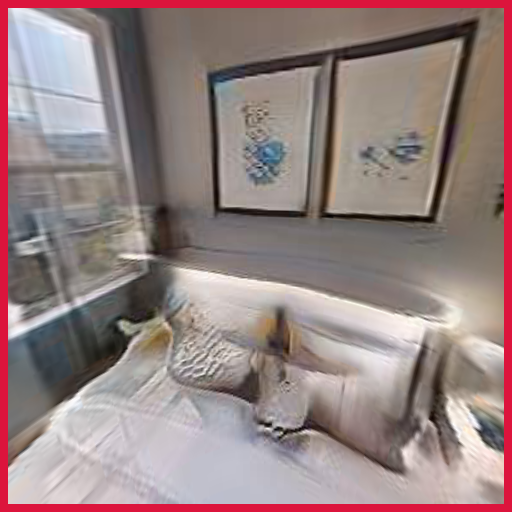}
  \end{subfigure}
  \hfill 
  \begin{subfigure}{0.48\columnwidth}
    \includegraphics[width=\textwidth]{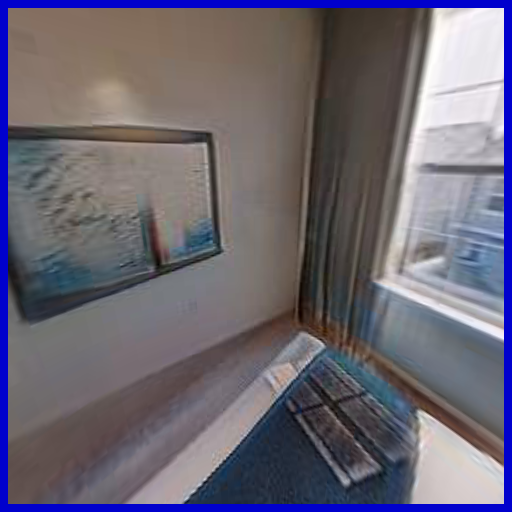}
  \end{subfigure}
  \end{minipage}
  \\[3mm]
  \begin{subfigure}{0.49\columnwidth}
    \includegraphics[width=\textwidth]{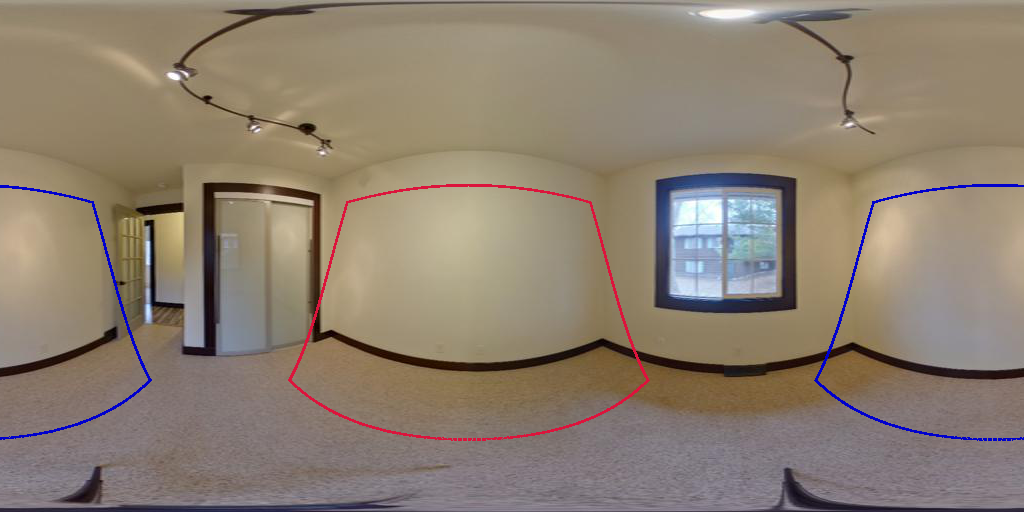}
  \end{subfigure}
  \hfill 
  \begin{subfigure}{0.49\columnwidth}
    \includegraphics[width=\textwidth]{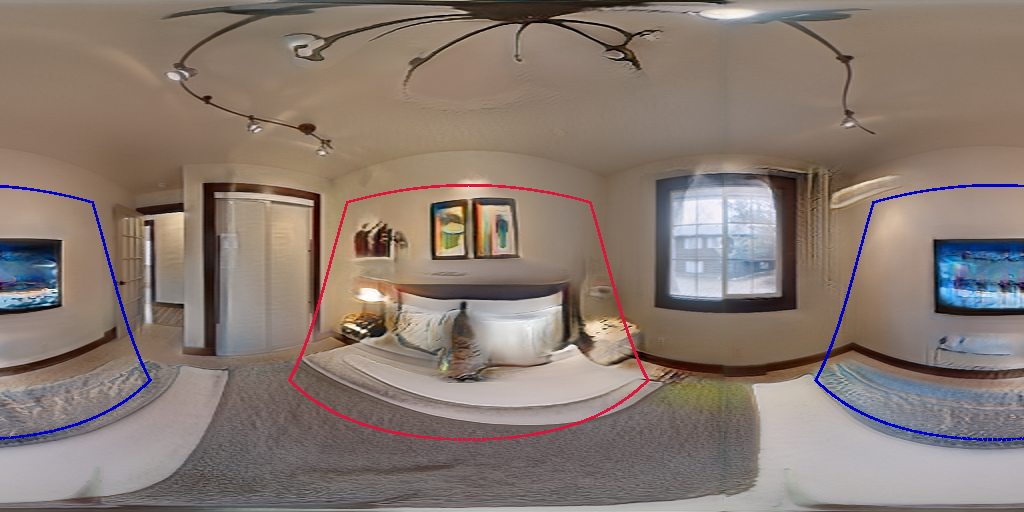}
  \end{subfigure}
  \\[0.8mm]
  \subfloat[Input]{\begin{minipage}[c]{0.49\columnwidth}
  \begin{subfigure}{0.48\columnwidth}
    \includegraphics[width=\textwidth]{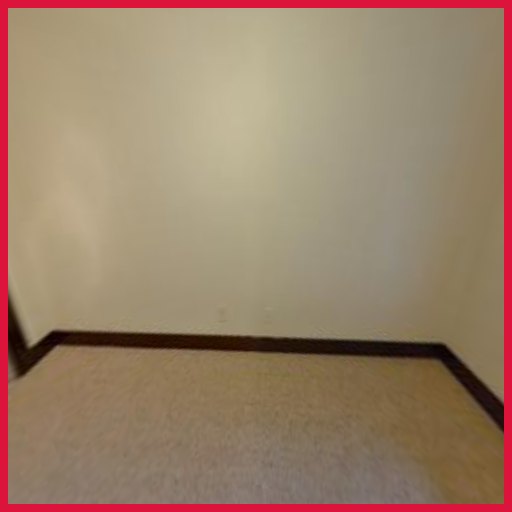}
  \end{subfigure}
  \hfill 
  \begin{subfigure}{0.48\columnwidth}
    \includegraphics[width=\textwidth]{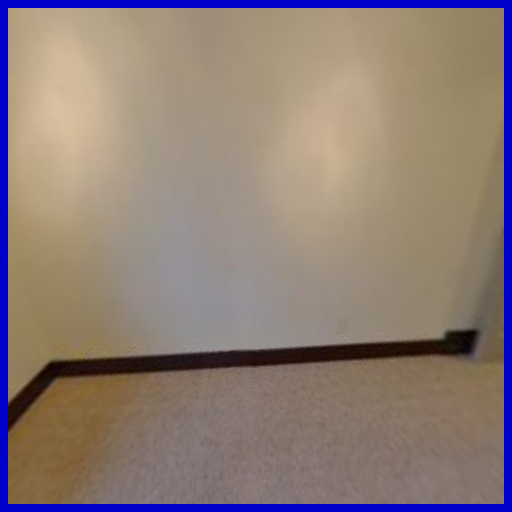}
  \end{subfigure}
  \end{minipage}}
  \hfill
  \subfloat[Generated]{\begin{minipage}[c]{0.49\columnwidth}
    \begin{subfigure}{0.48\columnwidth}
    \includegraphics[width=\textwidth]{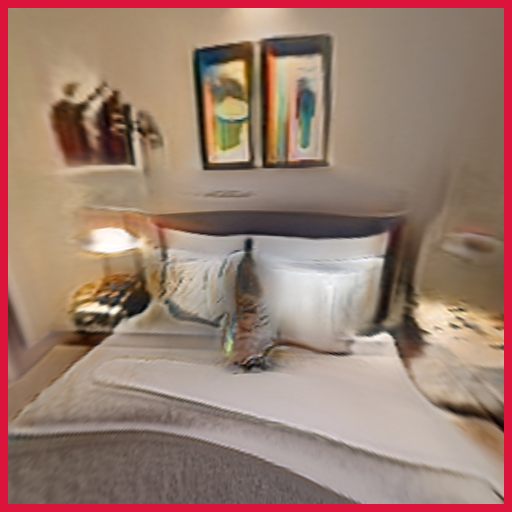}
    \end{subfigure}
    \hfill 
    \begin{subfigure}{0.48\columnwidth}
    \includegraphics[width=\textwidth]{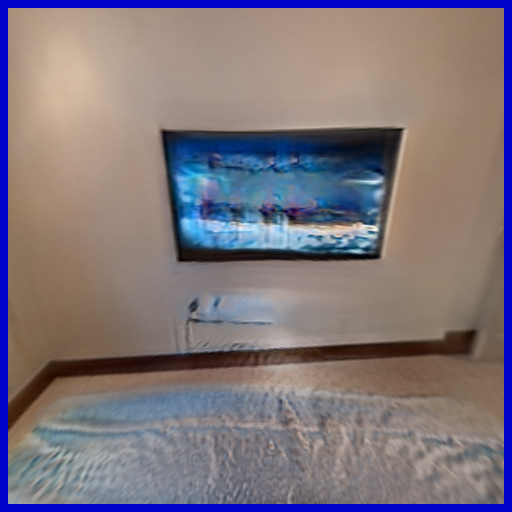}
  \end{subfigure}
  \end{minipage}}

\caption{Generation results of our method on real-world scenes. Input images are from the 360\degree bedrooms of the ZInD~\cite{ZInD}. More results are in the supplementary material.}
\label{fig:generalization}

\end{figure}

\noindent\textbf{Generalization to real-world images.}
We validate the generalization ability of our method on real-world scenes from the ZInD. As shown in Figure~\ref{fig:generalization}, our model generates plausible decorated images given real-world undecorated 360\degree images. Fine objects can also be generated to fit different bedroom structures.

To quantitatively evaluate the generalization quality, we run our model and all the I2I baselines on the ZInD. We do not include the layout-based methods in this experiment due to lack of ground-truth object labels. We evaluate all the methods using FID and KID scores on both the ZInD and the decorated split of the Structured3D dataset. The reported results in Table~\ref{tab:quantitative_results_zillow} show the superior generalization ability of our method over all the I2I baselines on real-world data. 


\begin{table}[]
    \centering
    \begin{tabular}{l|cc|cc}
        \toprule
         & \multicolumn{2}{c|}{vs. Structured3D} & \multicolumn{2}{c}{vs. ZInD} \\
        \midrule
        Method & $\mathrm{FID} \downarrow$ & $\mathrm{KID} \downarrow$ & $\mathrm{FID} \downarrow$ & $\mathrm{KID} \downarrow$ \\
        \midrule
        Pix2PixHD~\cite{wang2018high} & 114.70 & 71.94 & 93.29 & 75.30\\ 
        StarGANv2~\cite{choi2020stargan} & 93.56 & 49.47 & 73.73 & 58.80\\ 
        StyleD~\cite{kim2022style} & 102.02 & 70.36 & 59.43 & 45.78\\
        \midrule
        Ours & ~\textbf{88.86} & ~\textbf{47.29} & ~\textbf{51.56} & ~\textbf{33.74}\\ 
        \bottomrule
    \end{tabular}
    \caption{Quantitative evaluation on real-world images from the ZInD and the decorated split of the Structured3D dataset.}
    \label{tab:quantitative_results_zillow}
\end{table}

\subsection{Ablation studies}
\label{sec:ablation}

\noindent\textbf{Effectiveness of the conditional layout generator.} Recall that the layout generator creates a group of object vectors from an input image. These vectors are fused into an ellipse-like 360\degree object layout for further generation. To validate the layout generator, we disable it in our pipeline and simply pass the input image to the scene decorator. Furthermore, to validate the 360\degree setting for the object layout, we do not apply the 360\degree conversion in Eq.~(\ref{eq:distance:opacity_and_alphacomposition}) but rather fuse all raw pixels into a naive 2D layout (like in BlobGAN~\cite{epstein2022blobgan}). We report the results of this experiment in Table~\ref{tab:ablation_layout}, which clearly shows the effectiveness of our layout design.


\begin{table}[]
    \centering
    \begin{tabular}{l|cc}
        \toprule
        Method & $\mathrm{FID} \downarrow$ & $\mathrm{KID} \downarrow$  \\
        \midrule
        w/o layout generator & 99.64 & 72.33\\
        Traditional 2D layout~\cite{epstein2022blobgan} & 75.27 & 25.57 \\
        Ours (full pipeline) & \bf{69.17} & \bf{19.54}\\
        \bottomrule
    \end{tabular}
    \caption{Effectiveness of our proposed 360\degree object layout.}
    \label{tab:ablation_layout}
\end{table}

\noindent\textbf{Effectiveness of the pretrained emptier and cycle loss.} We show the effectiveness of the pretrained emptier and cycle loss in our pipeline in Table~\ref{tab:ablation_emptier}. Specifically, in this experiment, we disable the emptier and remove the cycle loss from the total loss. We also consider replacing the cycle consistency loss with a pairwise loss. We also validate the necessity of pretraining of the emptier. Table~\ref{tab:ablation_emptier} verifies the improvement gained by the pretraining and the cycle loss.


\begin{table}[]
    \centering
    \begin{tabular}{ll|cc}
        \toprule
        Emptier & Consistency loss & $\mathrm{FID} \downarrow$ & $\mathrm{KID} \downarrow$ \\
        \midrule
        w/o emptier & N/A & 91.28 & 34.40\\
        w/o emptier & pairwise loss & 73.40 & 23.77\\
        \midrule
        w/o pretraining & cycle loss & 76.74 & 23.67\\
        pretraining & cycle loss & \bf{69.17} & \bf{19.54}\\
        \bottomrule
    \end{tabular}
    \caption{Effectiveness of our proposed scene emptier.}
    \label{tab:ablation_emptier}
\end{table}

\subsection{User study}
\begin{figure}[t]
  \centering
    \includegraphics[width=\linewidth]{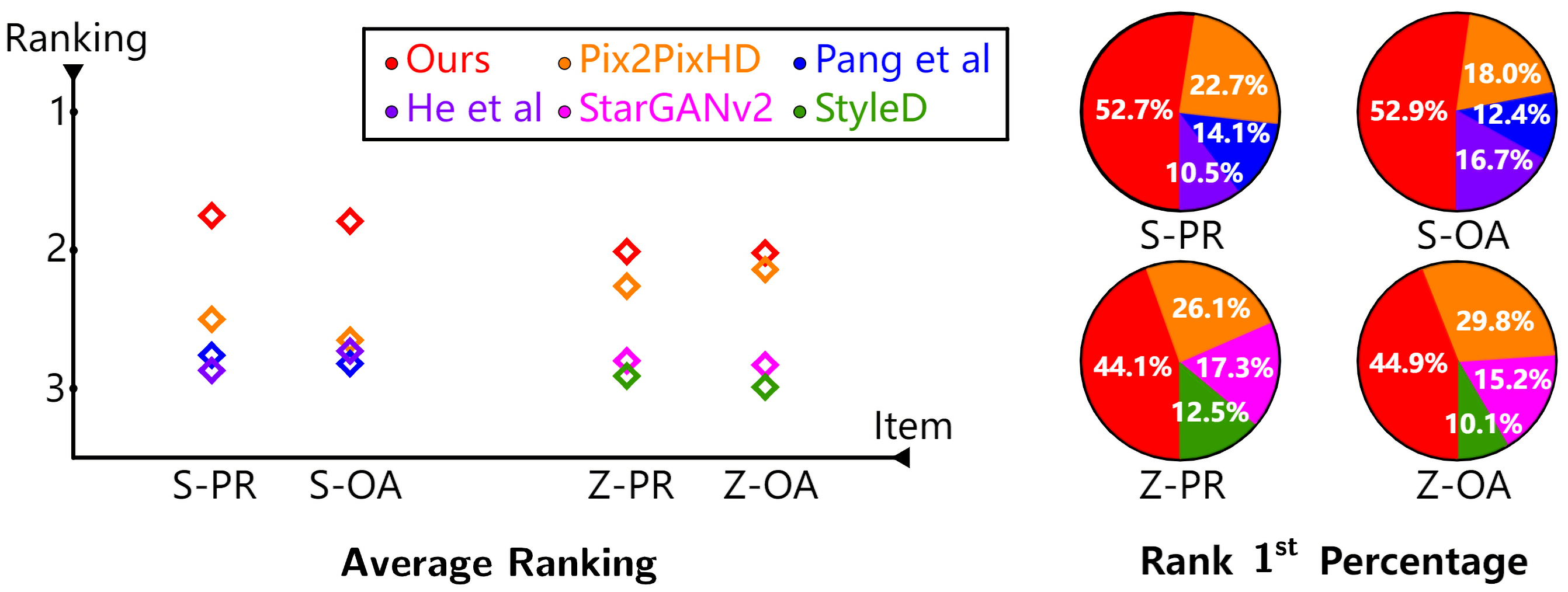}
    \caption{User study results. We show the average rankings and the percentage of the participants selecting the methods as ranking $1^{st}$. The symbol S and Z refer to the Structured3D and ZInD. PR and OA refer to photo-realism and object arrangement.}
    \label{fig:user_study}
\end{figure}

We conducted a user study on the generation quality of our method and other baselines. We presented generated 360\degree images in perspective views to participants. Since objects are often generated at the middle and bottom of output images, we randomly rendered two perspective views with the camera facing toward these areas. We took generation results on both the Structured3D and ZInD, then asked participants to rank the results in regard to image photo-realism and quality of furniture arrangement among our work and other baselines. For the Structured3D dataset, we evaluate Pix2PixHD~\cite{wang2018high}, Pang et al.~\cite{pang2022nsd}, and He et al.~\cite{he2021context}. For the ZInD, benchmark Pix2PixHD~\cite{wang2018high}, StarGANv2~\cite{choi2020stargan}, and StyleD~\cite{kim2022style}. Figure~\ref{fig:user_study} shows the results of the user study with 35 participants. It is clearly seen that our generated images are preferred by the participants for both the datasets and in terms of photo-realism and furniture arrangement.

\section{Conclusion}
\label{sec:Conclusion}
We propose a conditional image generative model to solve the task of automatic neural scene decoration in the 360\degree viewer. Our method offers immersive experiences of indoor scenes while enabling the controllability of generated content. We show that our method can generate realistic 360\degree images with diverse furniture layouts on the synthetic Structured3D dataset and generalize well to the real-world Zillow indoor dataset. As 360\degree images provide better context for scene understanding, an interesting research direction is to incorporate structural and semantic scene understanding into layout and image generation to improve furniture arrangement and object controllability. Our method also shares the limitations of generative models, i.e., the generation quality largely depends on the scale of the training dataset, which we aim to improve in our future work.


{\small
\bibliographystyle{ieee_fullname}
\bibliography{egbib}
}

\end{document}